\documentclass{IEEEtran}

\usepackage{amsbsy}
\usepackage{floatflt} 
\usepackage{amsthm}
\usepackage{tikz}
\usetikzlibrary{arrows.meta}
\usepackage{amsmath}
\usepackage{amssymb}
\usepackage{times}
\usepackage{graphics}
\usepackage{graphicx}
\usepackage{xspace}
\usepackage{paralist} 
\usepackage{setspace} 
\usepackage{xypic}
\xyoption{curve}
\usepackage{latexsym}
\usepackage{ifthen}
\usepackage{caption}
\usepackage{subfigure}
\usepackage{booktabs}
\usepackage{algorithm}
\usepackage{algorithmic}
\usepackage{color}
\usepackage{authblk}

\graphicspath{ {./images/} }

\newtheorem{theorem}{Theorem}
\newtheorem{remark}{Remark}

\newtheorem{assumption}{Assumption}

\newcommand{\R}{\mathbb{R}}
\DeclareMathOperator*{\argmin}{arg\,min}

\title{Personalized Federated Learning via Convex Clustering}
\author[1]{Aleksandar Armacki}
\author[2]{Dragana Bajovic}
\author[3]{Dusan Jakovetic}
\author[1]{Soummya Kar}

\affil[1]{Carnegie Mellon University, Pittsburgh, PA \authorcr Email: {\tt \{aarmacki, soummyak\}@andrew.cmu.edu}\vspace{1.5ex}}
\affil[2]{Faculty of Technical Sciences, University of Novi Sad, Novi Sad, Serbia \authorcr Email: {\tt dbajovic@uns.ac.rs} \vspace{1.5ex}}

\affil[3]{Faculty of Sciences, University of Novi Sad, Novi Sad, Serbia \authorcr Email: {\tt dusan.jakovetic@dmi.uns.ac.rs} \vspace{-2ex}}

\begin{document}

\maketitle

\begin{abstract}
We propose a parametric family of algorithms for personalized federated
learning with locally convex user costs. The proposed framework is based
on a generalization of convex clustering in which the differences between different users' models are penalized via a sum-of-norms penalty, weighted by a penalty parameter~$\lambda$. The proposed approach enables ``automatic" model clustering, without prior knowledge of the hidden cluster structure, nor the number of clusters. Analytical bounds on the weight parameter, that lead to simultaneous personalization, generalization and
automatic model clustering are provided. The solution to the formulated problem enables personalization, by providing different models across different clusters, and generalization, by providing models different than the per-user models computed in isolation. We then provide an efficient algorithm based on the Parallel Direction Method of Multipliers (PDMM) to solve the proposed  formulation in a federated server-users setting. Numerical experiments corroborate our findings. As an interesting byproduct, our results provide several generalizations to convex clustering.
\end{abstract}

\section{Introduction}

Federated learning (FL) is a paradigm in which many users collaborate, with the goal of learning a joint model \cite{pmlr-v54-mcmahan17a}. Each user has a local dataset, with private and possibly sensitive data. The data distribution across users is typically highly heterogeneous. \let\thefootnote\relax\footnote{\noindent The work of D. Bajovic is supported by the European Union’s Horizon 2020 Research and Innovation program under grant agreement No 957337. The work of A. Armacki and S. Kar was partially supported by the National Science Foundation under grant CNS-1837607. This paper reflects only the authors' views and the European Commission cannot be held responsible for any use which may be made of the information contained therein. }

A federated learning system can have a huge amount of users, wherein each user contributes with a proportionally small local dataset. Therefore, the federation provides users with the benefit of training on the joint data, effectively offering broader knowledge and better generalization. However, due to the highly heterogeneous nature of the data, it is nontrivial to design a FL system where individual users achieve better performance though the federation when compared with models trained on their own local data. In fact, the authors in~\cite{yu2020salvaging} show that in many tasks, users may actually not benefit from participating in federated learning.  The globally trained model underperforms on their local data, compared to the model solely trained on the local data. Moreover, applying privacy preserving techniques further deteriorates the performance. On the other hand, users with very small datasets suffer from overfitting and poor generalization of models trained only on their local data.

To amend these problems and reap the benefits of both worlds -- the abundance of data and better learning that the federation offers, as well as adapting the models to perform well on the local data, \textit{personalized federated learning} is introduced. Unlike the standard federated learning, the goal of personalized federated learning is to learn multiple models. In particular, let $N$ be the number of participating users, with $f: \mathbb{R}^d \mapsto \mathbb{R}$ a given cost function. Then, the goal of standard FL is to solve
\begin{equation}\label{eq:std_fed}
    \argmin_{x \in \mathbb{R}^d}F_{\text{global}}(x) = \frac{1}{N}\sum_{i = 1}^Nf_i(x), 
\end{equation} where $f_i(x)$ is the cost function $f$ evaluated on the local dataset of the $i$-th user. Contrary to this approach, the (broad) goal of personalized FL is to learn $N$ models, by solving
\begin{equation}\label{eq:per_fed}
    \argmin_{x_1,\ldots,x_N \in \mathbb{R}^d} F_{\text{local}}(x_1,\ldots,x_N) = \frac{1}{N}\sum_{i = 1}^Nf_i(x_i),
\end{equation} subject to appropriately defined constraints. Depending on the constraints imposed and the approach taken to solving (\ref{eq:per_fed}), the literature on personalized FL adopts different approaches to personalization, including multi-task learning \cite{mocha}, \cite{richtarik}, fine-tuning \cite{fallah_meta}, \cite{wang2019federated}, knowledge distillation \cite{hinton}, \cite{zhang2018deep}, \cite{Bistritz2020DistributedDF}, \cite{joshi} and clustering-based approaches \cite{ghosh2021efficient}, \cite{sattler}, \cite{Mansour20}.

In this paper, we propose a novel approach to personalized federated learning that enables simultaneous personalization, generalization, and model clustering. The approach is based on the following novel personalized FL (convex problem) formulation:
\begin{equation}\label{eq:per_fed_cc}
   \argmin_{x_1,\ldots,x_N \in \mathbb{R}^d} F(x_1,\ldots,x_N) = \frac{1}{N}\sum_{i = 1}^Nf_i(x_i) + \lambda\sum_{j\neq i} \|x_i-x_j\|,
\end{equation}
where $\lambda>0$ is a penalty parameter, and $\| \cdot \|$ stands for the Euclidean norm. 
Compared with~\eqref{eq:per_fed}, the formulation~\eqref{eq:per_fed_cc} has a regularization term controlled with~$\lambda>0$,  that penalizes the differences between the local nodes' solutions via a sum-of-norms penalty.  

Formulation~\eqref{eq:per_fed_cc} may be seen  as a generalization of convex clustering, e.g.,~\cite{Sun21}, where, instead of the (quadratic) distance functions, which penalize the departure from a local data point, we use arbitrary convex local losses. 

Problem~\eqref{eq:per_fed_cc} is also  related with the personalized FL formulation  in~\cite{Hanzely20}, which, instead of the sum of norms of pairwise distances uses the sum of \emph{squared} norms of distances of local solutions to their average. As we show in the paper, although the two formulations are resembling, the solution structures of the two formulations are qualitatively very different.   

Specifically, as we show in the paper, the solution to~\eqref{eq:per_fed_cc} has several interesting properties, which, to the best of our knowledge, are not (jointly) exhibited with any of the previous personalized FL  formulations. Namely, a  solution to~\eqref{eq:per_fed_cc} has a clustered structure: depending on the penalty parameter $\lambda$ and similarity of the local functions, local solutions $x_i^\star$ of (\ref{eq:per_fed_cc}) are equal across certain users' groups (clusters). The number of groups (clusters) $K$ is automatically determined as part of the solution. To further illustrate benefits of this feature, suppose that the users exhibit an (unknown) clustered structure such that each user’s data within a cluster comes from the same distribution, while the distributions that correspond to different clusters are mutually different. If the different clusters' distributions are sufficiently far apart, the proposed method (\ref{eq:per_fed_cc}) uncovers the unknown cluster structure and hence allocates the same models to all users within the same cluster. This allows within-cluster generalization, i.e., users to effectively enlarge their training data by harnessing data of all users from the same cluster. In addition, depending on the distance between the different clusters' distributions, the method (\ref{eq:per_fed_cc}) allows for a controlled across-clusters generalization; intuitively, it allows a user to harness data from another clusters' distribution, but with a different (reduced) "weight" when compared to within-cluster data. If, at an extreme, the difference between different clusters' distributions is negligible (but this information is unknown), then users should clearly use the global model (\ref{eq:std_fed}). This is again captured by (\ref{eq:per_fed_cc}), because (as shown in the paper) it matches (\ref{eq:std_fed}) for $\lambda$ above a threshold.

In summary, our contributions are the following: First, we propose a novel formulation for personalized federated learning~\eqref{eq:per_fed_cc}, the solution of which has a clustering structure while at the same time preserving generalization abilities. 

Second, we provide a condition on the penalty parameter $\lambda$, with theoretical guarantees, for discovering the ``hidden structure'' underlying the models; this condition is expressed in terms of the well-established diversity of the local functions, hence making a strong connection and justifying analytically the use of this quantity. 

Third, the proposed solution "automatically" determines the number of models $K$, i.e., $K$  need not be known in advance.

Fourth, we provide an efficient algorithm to solve the novel personalized learning formulation~\eqref{eq:per_fed_cc} in a federated server-client setting that is based upon the Parallel Direction Method of Multipliers (PDMM)~\cite{TomLuoPDMM}. 

Finally, we demonstrate by simulation examples, on a supervized binary classification problem, that the proposed solution exhibits $1)$ generalization, i.e., improves testing accuracy with respect to the users models trained in isolation; $2)$ personalization, i.e., improves testing accuracy with respect to the global FL model (\ref{eq:std_fed}); and $3)$ achieves a comparable (or better) generalization and personalization (in the sense of $1)$ and $2)$) than \cite{richtarik}, while at the same time uncovering cluster structure, hence reducing the number of distinct models from $N$ to $K$.   

With respect to existing personalized FL approaches discussed above, the works \cite{ghosh2021efficient,sattler,Mansour20,joshi} 
 also account for users' clustering in a certain way, but very differently from our approach. Most notably, existing approaches aim to uncover ``cluster identities'' first and subsequently
provide loss minimizations across cluster groups in isolation from other groups. This within-clusters isolation may reduce overall 
generalization ability of the models. In contrast, the proposed approach allows also for across-clusters generalization that is further controlled by the penalty parameter~$\lambda$. It is worth noting that reference \cite{BianchiRobust} introduces formulation similar to (\ref{eq:per_fed_cc}), but in a different context of distributed consensus optimization. Most importantly, they are only concerned with the question when (\ref{eq:per_fed_cc}) matches (\ref{eq:std_fed}), i.e., when (\ref{eq:per_fed_cc}) leads to a global consensus across local models; they are not concerned, nor they study personalization (clustering) abilities of (\ref{eq:per_fed_cc}).

Our results are also of direct interest to convex clustering, e.g.,~\cite{Sun21}, as they
 provide recovery guarantees for generalized convex clustering, when the squared quadratic loss $f_i(x):=\|x_i-a_i\|^2$ per data point~$a_i \in {\mathbb R}^d$ is replaced with an arbitrary differentiable convex loss, e.g., the Huber loss.

\textbf{Paper organization}. The rest of the paper is organized as follows. Section \ref{sec:problem_formulation} describes the problem of interest and outlines the assumptions used in the analysis. Section \ref{sec:theoretical_guarantees} presents the recovery guarantees of the method. Section \ref{sec:proposed_algo} outlines an efficient algorithm for solving the proposed problem in the federated setting. Section \ref{sec:numerical_results} presents numerical experiments, and Section \ref{sec:conclusion} concludes the paper. The next paragraph introduces the notation used throughout the paper.

\textbf{Notation}. The set of real numbers is denoted by $\R$, while $\R^d$ denotes the corresponding $d$-dimensional vector space; $\| \cdot \|$ represents the standard Euclidean norm. $\langle \cdot, \cdot \rangle$ represents the standard vector product over the space of real vectors. $[N]$ denotes the set of integers up to and including $N$, i.e., $[N] = \{1,\ldots,N\}$.

\section{Problem formulation}\label{sec:problem_formulation}

Consider a collection of $N$ users, $i=1,\ldots,N$, that %
participate in a federated learning activity. 
Each user $i$ holds a function $f_i:\,{\mathbb R}^d \rightarrow {\mathbb R}$. Function $f_i$ may correspond, e.g., to an empirical loss with respect to the local data set available at user~$i$. We make the following assumptions throughout the paper. 

\begin{assumption}
\label{assumption-f-i-s}
For each $i=1,\ldots,N$, function 
$f_i:\,{\mathbb R}^d \rightarrow {\mathbb R}$ 
is convex and coercive, i.e., 
$f_i(x) \rightarrow +\infty$ whenever $\|x\| \rightarrow +\infty$.
\end{assumption}

\begin{assumption}
\label{assumption-Lip-grad}
For each $i=1,\ldots,N$, function $f_i:\,{\mathbb R}^d \rightarrow {\mathbb R}$ 
has Lipschitz continuous gradients, i.e. the following holds
\begin{equation*}
    \| \nabla f_i(x) - \nabla f_i(y) \| \leq L\|x - y\|, \: \mathrm{for\,all\,} \: x,y \in {\mathbb R}^d.
\end{equation*} 
\end{assumption} Note that, under the above assumptions, 
problems~\eqref{eq:std_fed} and~\eqref{eq:per_fed_cc} are solvable. We denote by $y^\star \in {\mathbb R}^d$ a solution to~\eqref{eq:std_fed} and by 
$\{x_i^\star\}$, $i=1,\ldots,N$, 
$x_i^\star \in {\mathbb R}^d$, a solution to~\eqref{eq:per_fed_cc}.

There are many machine learning models that satisfy Assumptions~\ref{assumption-f-i-s} and~\ref{assumption-Lip-grad}, such as supervized binary classification problems studied in Section \ref{sec:numerical_results}.

The high-level goal in personalized federated learning is that each user $i$ finds a local model, say $x_i^{\bullet} \in {\mathbb R}^d$, 
that performs well on the local data (i.e., the value $f_i(x_i^{\bullet})$ is low), but that also 
exhibits a generalization ability with respect to data available at other users~$j \neq i$. 
In addition, a desirable feature of personalized federated learning is that the users are able to  classify other users into two categories. The first  category corresponds to those users $j \in \{1,2,\ldots,N\}$ that have similar data (similar $f_j$'s) to their own; the second category corresponds to those users whose local data is ``sufficiently different'' from theirs. With this classification in place, each user $i$ can fully harness the data from ``similar users'' for 
an improved personalization while avoiding overfitting; e.g., when user~$i$ has a very few data points of its own, it effectively enlarges its data set while preserving personalization. On the other hand, the data from ``sufficiently different users'' should still be harnessed in a controlled way to further improve generalization abilities. 

To account for the effects above, we provide a novel personalized learning formulation, where 
each user $i$ wants to obtain the local model 
$x_i^\star \in {\mathbb R}^d$ 
such that $x^\star:=((x_1^\star)^\top,\ldots,(x_N^\star)^\top )^\top \in {\mathbb R}^{N\,d}$  
is a minimizer of~\eqref{eq:per_fed_cc}, where 
$\lambda>0$ is a tuning parameter. 
Intuitively, the term $\sum_{i=1}^N f_i(x_i)$ 
forces the local models $x_i$'s 
to behave well with respect to local costs $f_i$'s; 
the term $\lambda \, \sum_{j \neq i}\|x_i-x_j\|$
 makes the local models be mutually close, 
 hence enabling generalization.
   The penalization term $\sum_{j \neq i} \|x_i-x_j\| $ is known to enforce sparsity in other contexts, in the sense that 
   it forces many of the $x_i$'s to be mutually equal 
   at a solution of \eqref{eq:per_fed_cc}.

It is interesting to compare our novel formulation \eqref{eq:per_fed_cc} with 
the personalized federated learning formulation in \cite{richtarik}:
\begin{equation}\label{eq:per_fed_richtarik}
   \argmin_{x_1,\ldots,x_N \in \R^d}  \frac{1}{N}\sum_{i = 1}^Nf_i(x_i) + \gamma\sum_{j\neq i} \|x_i-x_j\|^2, 
\end{equation} where $\gamma > 0$ is a penalty parameter. 

 The difference of \eqref{eq:per_fed_richtarik} with respect to 
 \eqref{eq:per_fed_cc} is that, in \eqref{eq:per_fed_richtarik}, the differences of local models $x_i$ and $x_j$ are penalized via the squared Euclidean norm, while with 
 our formulation, the 2-norm appears without squares. 
 There are several important implications of this difference with respect to the resulting personalized learning models. Most importantly, in contrast with \eqref{eq:per_fed_cc},   formulation \eqref{eq:per_fed_richtarik} in general  
 does not lead to model clustering for any $\lambda>0$. In addition, as a side comment, the solutions 
 to \eqref{eq:std_fed} and \eqref{eq:per_fed_richtarik} are in general mutually different for any $\lambda > 0$
  (in the sense that the solution $\{y_i^\star\}$ to 
  \eqref{eq:per_fed_richtarik} does not obey $y_i^\star=y_j^\star$, for all $i,j$, irrespective of the choice of $\lambda$). In contrast, with \eqref{eq:per_fed_cc}, 
  we recover global model learning as in \eqref{eq:std_fed} for $\lambda > \widehat{\lambda}$\footnote{This can be easily seen based on Theorem 1 in \cite{BianchiRobust}.}. 
 
We also connect \eqref{eq:per_fed_cc} with convex clustering. Convex clustering, e.g.~\cite{Sun21}, is an appealing method to cluster $N$ data points $a_i \in {\mathbb R}^d$, $i=1,\ldots,N$. The method corresponds to solving problem \eqref{eq:per_fed_cc} 
with $f_i(x) = \frac{1}{2}\|x-a_i\|^2$. Intuitively, to each data point $i$, we associate a candidate cluster center $x_i$, and then we enforce a (soft) constraint that many $x_i$'s should be mutually equal. 
There are several 
efficient algorithms and cluster recovery 
guarantees results available for convex clustering, but only when $f_i(x) = \frac{1}{2}\|x-a_i\|^2$. Our results make a direct generalization of convex clustering to other loss metrics, e.g., the ``distance'' of a candidate cluster $x_i$ from data point $a_i$ may be measured through the Huber loss.

\section{Theoretical guarantees for optimal cluster recovery}\label{sec:theoretical_guarantees}

In this section, we state and prove 
our main results on characterization of solutions to~\eqref{eq:per_fed_cc}. 

We start by defining the following auxiliary optimization problem associated to a certain (predefined) partition of users $C_1,\ldots, C_K$, $\cup_{k=1}^K C_k= [N]$ and $C_k\cap C_j=\emptyset$: 
\begin{equation}\label{eq:clustered-cc}
   \argmin_{w_1,\ldots,w_N \in \R^d}  \frac{1}{N}\sum_{k = 1}^K n_k g_k(w_k) + \lambda\sum_{l\ne k}  n_k n_l\|w_k-w_l\|, 
\end{equation}
where $g_k(w):=1/n_k \sum_{i\in C_k} f_i(w)$, for $w\in \mathbb R^d$, and $n_k=|C_k|$ is the number of elements in $C_k$, for $k=1,\ldots,K$. Let $w_k^\star= w_k^\star(\lambda)$,  $k=1,\ldots,K$, denote a solution to~\eqref{eq:clustered-cc}. Note that problem~\eqref{eq:clustered-cc} is solvable by Assumption~\ref{assumption-f-i-s}. 

\begin{theorem}
\label{lemma-recovery-cond}
Consider problem~\eqref{eq:per_fed_cc}. Assume that, for some node partition $C_1, C_2,\ldots, C_K$, and parameter $\lambda$, there holds \begin{equation}
\label{eq:condition-recovery}
\lambda \geq \max_{k=1,\ldots,K} \max_{i,j\in C_k}\frac{ \|\nabla f_i(w_k^\star) - \nabla f_j (w_k^\star)\|}{n_k}.    
\end{equation}
Next, let $\{x_i^\star\}$, $i=1,\ldots,N$, 
be defined as follows: for each $i\in C_k$, we let  $x_i^\star = w_k^\star$, for $k=1,\ldots,K$, where  $\{w_k^\star=w_k^\star(\lambda)\}$, $k=1,\ldots,K$, is a solution to~\eqref{eq:clustered-cc}, defined for the same partition $C_1,\ldots,C_K$ that verifies~\eqref{eq:condition-recovery}. 
Then, $\{x_i^\star\}$, $i=1,\ldots,N$, is a solution of~\eqref{eq:per_fed_cc}.
\end{theorem}

\begin{remark} Note that Theorem~\ref{lemma-recovery-cond} guarantees that 
at least one solution of~\eqref{eq:per_fed_cc} 
exhibits the clustered structure with respect to partition~$C_1,\ldots,C_K$, while it does not preclude a scenario that there might be another solution of~\eqref{eq:per_fed_cc} that may not exhibit this cluster structure. However, when each of the $f_i$'s is in addition assumed to be strictly convex, then $\{x_i^\star\}$ is unique,  and it necessarily has the clustered structure.  
\end{remark}

We next prove Theorem~\ref{lemma-recovery-cond}. 
\begin{proof}
The proof is in spirit similar to Theorem 1 in~\cite{Panahi17}. From the first order optimality conditions for~\eqref{eq:clustered-cc}, we obtain that, for each $k=1,\ldots,K$, there must hold:
\begin{equation}
\label{eq:KKT-clustered-cc}
\nabla g_k(w_k^\star) +\lambda  \sum_{l\ne k} n_l r^\star_{kl}=0,\end{equation}
where $r^\star_{kl}$ is a subgradient of $\|w_k-w_l\|$ with respect to $w_k$, computed at the solution. For each $k,l=1,\ldots,K$, $r^\star_{kl}$ must satisfy:
\begin{equation}
\label{eq:subgrad-clustered-cc}
r^\star_{kl}= \left\{ \begin{array}{cc}
\frac{w_k^\star-w_l^\star}{\|w_k^\star-w_l^\star\|}     , &\mathrm{for\,}w_k^\star \neq w_l^\star  \\
\mathrm{a\, vector\,}r \in \mathbb R^d\mathrm{\,s.t.\,} \|r\|\leq 1,      & \mathrm{otherwise}
\end{array}\right.    
\end{equation}

We now turn to first order optimality conditions for the original problem~\eqref{eq:per_fed_cc}:
\begin{equation}
\label{eq:KKT-per-fed-cc}
\nabla f_i(x_i) + \lambda \sum_{j\neq i} s_{ij}=0,    
\end{equation}
where $s_{ij}$ is a subgradient of $\|x_i-x_j\|$ computed with respect to $x_i$. Similarly as in the above, at the solution, $s_{ij}$ must satisfy:
\begin{equation}
\label{eq:subgrad-per-fed-cc}
s_{ij}= \left\{ \begin{array}{cc}
\frac{x_i-x_j}{\|x_i-x_j\|}     , &\mathrm{for\,}x_i \neq x_j  \\
\mathrm{a\, vector\,}s \in \mathbb R^d\mathrm{\,s.t.\,} \|s\|\leq 1,      & \mathrm{otherwise}
\end{array}\right.    
\end{equation}
It can be verified that, when condition~\eqref{eq:condition-recovery} is fulfilled, then the following choice of $x_i^\star$ and $s_{ij}^\star$ satisfy the first order optimality conditions in~\eqref{eq:KKT-per-fed-cc} and~\eqref{eq:subgrad-per-fed-cc}
\begin{align}
x_i^\star &=w_k^\star, \,i\in C_k,\\    
s_{ij}^\star & = \left\{
\begin{array}{cc}
r^\star_{kl},     &  j\in C_l, k\neq l  \\
\frac{\nabla f_j(w_k^\star) - \nabla f_i (w_k^\star)}{\lambda n_k},     &  j\in C_k \\
\end{array}
\right.,
\end{align}
    hence proving the result.  
\end{proof}

Theorem \ref{lemma-recovery-cond} guarantees the existence of a solution of (\ref{eq:per_fed_cc}) that exhibits the desired clustering structure. However, if the parameter $\lambda$ is chosen too large, it can actually coarsen the clustering structure $C_1, \ldots, C_K$ and provide a solution with $1 \leq M < K$ groups (clusters). The following theorem ensure the correct clustering structure is recovered.

\begin{theorem}
\label{lemma-recovery-uperb}
Consider problem~\eqref{eq:clustered-cc}. If for some node partition $C_1, C_2,\ldots, C_K$ and parameter $\lambda>0$ there holds \begin{equation}
\label{eq:condition-unique}
\lambda < \frac{\min_{k,l \in [K], k\ne l} \|\nabla g_k(w_k^\star) - \nabla g_l (w_k^\star)\|}{2\max_{k \in [K]}\sum_{l \ne k}n_l},    
\end{equation}
then for each $k, l \in [K], k \ne l$, we have $w_k^\star \ne w_l^\star$, where $w_k^\star=w_k^\star(\lambda)$, $k=1,\ldots,K$, is a solution to~\eqref{eq:clustered-cc}, defined for the same partition $C_1,\ldots,C_K$, that verifies~\eqref{eq:condition-unique}.
\end{theorem}

We note that in practice, the bounds (\ref{eq:condition-recovery}) and (\ref{eq:condition-unique}) might not be easy to obtain, as $w^\star_k$'s depend on $\lambda$. In Appendix \ref{app:further_discuss} we provide several considerations regarding selection of the penalty parameter $\lambda$ in \eqref{eq:per_fed_cc}, in practice.
        
We next prove Theorem~\ref{lemma-recovery-uperb}. 
\begin{proof}
Denote by $s_k = \lambda \sum_{l \ne k}r^*_{kl}$. From~\eqref{eq:KKT-clustered-cc}, we have
\begin{align*}
    \|s_l - s_k \| &= \|\nabla g_k(w^*_k) - \nabla g_l(w^*_l)\| \\ &\leq \| \nabla g_k(w^*_k) - g_l(w^*_k) \| + \| \nabla g_l(w^*_k) - \nabla g_l(w^*_l)\| \\ &\leq \| \nabla g_k(w^*_k) - g_l(w^*_k) \| + L\| w^*_k - w^*_l \|, 
\end{align*} where we used Assumption~\ref{assumption-Lip-grad} in the second inequality. Rearranging, we get
\begin{equation*}
    \|w^*_k - w^*_l\| \geq \frac{1}{L}\| \nabla g_k(w^*_k) - g_l(w^*_k) \| - \frac{1}{L}(\|s_l\| + \|s_k \|).
\end{equation*} Next, note that
\begin{equation*}
    \|s_k\| \leq \lambda \sum_{l \ne k}n_l\|r^*_{kl}\| \leq \lambda \sum_{l \ne k}n_l. 
\end{equation*} Plugging in the equation above, we get
\begin{align*}
    \|w^*_k - w^*_l\| \geq \frac{1}{L}\| \nabla g_k(w^*_k) - g_l(w^*_k) \| - \frac{2\lambda}{L}\max_{k \in [K]}\sum_{l \ne k}n_l.
\end{align*} It can be readily checked that the choice of $\lambda$ satisfying~\eqref{eq:condition-unique} results in
\begin{equation*}
    \|w^*_k - w^*_l\| > 0, \quad k \ne l,
\end{equation*} hence showing the claim.
\end{proof}

\section{Algorithm for personalized federated learning}\label{sec:proposed_algo}

In this section, we introduce an algorithm to solve \eqref{eq:per_fed_cc} in a federated server-users setting. The algorithm is adapted from the parallel direction method of multipliers (PDMM) in~\cite{TomLuoPDMM}. 

We start by reformulating problem \eqref{eq:per_fed_cc} as follows:
\begin{eqnarray}
    \label{eqn-reformulation-personalized}
    &\,& \mathrm{min}\,\sum_{i=1}^N f_i(x_i)
    + \lambda \sum_{j \neq i} \|z_{ij}\| \\
    &\,&\mathrm{s.t.}\,\,x_i-x_j=z_{ij},\,\,i\neq j. \nonumber 
\end{eqnarray}
That is, each of the $N(N-1)$ terms 
$\|x_i-x_j\|$ in \eqref{eq:per_fed_cc}
 are replaced with $\|z_{ij}\|$, where $z_{ij}
 \in {\mathbb R}^d$ is  
an auxiliary (primal) variable. Then, for equivalence of \eqref{eq:per_fed_cc} and \eqref{eqn-reformulation-personalized},
 we add for each ordered pair $(i,j)$, $i \neq j$,
 the constraint $x_i-x_j=z_{ij}$. 
 Next, introduce the augmented Lagrangian 
 $L:\, {\mathbb R}^{N\,d} 
 \times {\mathbb R}^{N(N-1)d} \times 
 {\mathbb R}^{N(N-1)d} \rightarrow \mathbb R$, 
 defined by:
\begin{align}
\begin{aligned}
\label{eqn-AL}
&\,& L(\{x_i\},\{z_{ij}\}, \{\mu_{ij}\}) =
\sum_{i=1}^N {f}_i(x_i) + \lambda\,\sum_{j \neq i}\|z_{ij}\|
\\
&+& \sum_{j \neq i}
\mu_{ij}^\top \left( x_i-x_j-z_{ij} \right)
+ \frac{\rho}{2}
\sum_{j \neq i}\|x_i-x_j-z_{ij}\|^2,
\end{aligned}
\end{align}
where $\{x_i\}$, $i=1,\ldots,N$, and $\{z_{ij}\}$, 
$i=1,\ldots,N$, $j=1,\ldots,N$, $i \neq j$, are the primal variables, and $\{\mu_{ij}\}$, 
$i=1,\ldots,N$, $j=1,\ldots,N$, $i \neq j$, are the dual variables, and $\rho>0$ is a penalty parameter.\footnote{It is possible to halve the number of constraints in \eqref{eqn-reformulation-personalized} 
 by imposing the constraint $x_i-x_j=z_{ij}$ only for $i<j$. This approach reduces the number of variables at the cost of additional coordination of users on the server's side. We  present here the approach with the larger number of variables and less coordination required.}  
Abstracting details, PDMM proceeds as follows. First, it updates at each iteration  $t=0,1,\ldots$, a randomly selected subset of 
primal variables $\{x_i,z_{ij}\}$ by minimizing a surrogate of $L$ with the rest of primal and dual variables fixed. Then, a randomly selected subset of dual variables $\{\mu_{ij}\}$ is updated, while also bookkeeping a set of auxiliary dual variables $\{\widehat{\mu}_{ij}\}$. See equations (29)--(31) in \cite{TomLuo1} for a detailed definition of the generic PDMM. 

Here, we apply and adapt PDMM to solve  \eqref{eqn-reformulation-personalized}, and hence, solve \eqref{eq:per_fed_cc}, in the federated server-users setting. 
To facilitate presentation of the algorithm, we enumerate all primal variables $x_i$'s and $z_{ij}$'s through a common index set $\mathcal{S}_P$ with $N^2$ elements, such 
that the $i$-th element of $\mathcal{S}_P$, $i=1,\ldots,N$, corresponds to $x_i$, and 
the remaining $N(N-1)$ subsequent elements correspond to $z_{ij}$'s, where the ordered pairs $\ell \sim (i,j)$ are positioned  lexicographically in $\mathcal{S}_P$. 
For example, $(N+1)$-th element of $\mathcal{S}_P$ corresponds to 
 variable $z_{12}$, $(N+2)$-nd element of 
 $\mathcal{S}_P$ corresponds to $z_{13}$, etc. 
Similarly, 
we let $\mathcal{S}_D$ be the $N(N-1)$-sized index set, such that its $\ell$-th element corresponds to 
the dual variable $\mu_{ij}$, $\ell \sim (i,j)$, 
$\ell=1,\ldots,N(N-1)$.
The PDMM-based personalized FL method is shown in Algorithm~\ref{alg-PDMM-FL}. 

\begin{algorithm}[htp]
\begin{center}
    \begin{algorithmic}
	\caption{A PDMM-based algorithm for personalized FL and model clustering}
	\label{alg-PDMM-FL}
	\FOR  {$t=0, \ldots, T-1$}
		\STATE (S1) The server randomly selects a subset ${\mathcal S}_P^{(t)} \subset \mathcal{S}_P$ of 
		$S_P$, $S_P<N^2$, primal variables;
		\STATE (S2) Each user $i \in \{1,\ldots,N\}$, such that $i \in {\mathcal S}_P^{(t)}$, 
		performs the update of $x_i^{(t)}$ as follows:
		 \begin{eqnarray}
&\,& x_i^{(t+1)} = 
		\mathrm{arg\,min}_{x_i \in {\mathbb R}^d} \,
		f_i(x_i)+ 
		\sum_{j \neq i} 
		(\widehat{\mu}_{ij}^{(t)})^\top x_i
		\nonumber \\
		 \label{eqn-x-i-update-PDMM-1}
		&+&
		\frac{\rho}{2}\sum_{j \neq i}
		\|x_i-x_j^{(t)}-z_{ij}^{(t)}\|^2\\
		&+& \nonumber
		\eta_i\,B_i(x_i,x_i^{(t)});
\end{eqnarray}
		\STATE (S3) Each user $i \in \{1,\ldots,N\}$,  such that $\ell \in {\mathcal S}_P^{(t)}$, $\ell \sim (i,j)$, 
		performs the update of $z_{ij}^{(t)}$ as follows:
\begin{eqnarray}
&\,& z_{ij}^{(t+1)} = 
		\mathrm{arg\,min}_{z_{ij} \in {\mathbb R}^d} \,
		\lambda\,\|z_{ij}\|-
		(\widehat{\mu}_{ij}^{(t)})^\top
		z_{ij} \nonumber \\
		&+&
		 \label{eqn-z-ij-update-PDMM-1}
		\frac{\rho}{2}
		\|x_i^{(t)}-x_j^{(t)}-z_{ij}\|^2\\ 
		&+& \nonumber
		\eta_{ij}\,B_{ij}(z_{ij},z_{ij}^{(t)});
		\end{eqnarray}
		\STATE (S4) 
		For each $s \in \{1,\ldots,N\}$, and  
		$\ell$, $\ell \sim (i,j)$, such 
		that $s, \ell \notin \mathcal{S}_P^{(t)}$, 
		set 
		$x_s^{(t+1)}=x_{s}^{(t)}$, 
		and $z_{ij}^{(t+1)}=z_{ij}^{(t)}$;
		\STATE (S5) Each user 
		$i\in \{1,\ldots,N\}$, such that $i \in \mathcal{S}_P^{(t)}$, sends $x_i^{(t+1)}$ to the server;
		\STATE (S6) Each user 
		$i \in \{1,\ldots,N\}$, such that 
		$\ell \in \mathcal{S}_P^{(t)}$, 
		$\ell \sim (i,j)$, sends $z_{ij}^{(t+1)}$ to the server;
		\STATE (S7) The server collects 
		$\{x_i^{(t+1)}\}$, $i \in \mathcal{S}_P^{(t)}$, 
		and 
		broadcasts this $(S_P\,d) \times 1$ vector to all users $i \notin \mathcal{S}_P^{(t)}$;
		\STATE (S8) The server picks a random subset $\mathcal{S}_D^{(t)}$, 
		$\mathcal{S}_D^{(t)}
		\subset \mathcal{S}_D$, of $S_D$
		dual variables, and performs 
		the following update for $\ell 
		\in \mathcal{S}_D$, $\ell \sim (i,j)$:
		\begin{eqnarray}
		\label{eqn-mu-ij-PDMM}
		&\,& \mu_{ij}^{(t+1)} = \mu_{ij}^{(t)}\\ 
		&+& \nonumber
		\tau\,\rho\,\left( 
x_i^{(t+1)}-x_j^{(t+1)}-z_{ij}^{(t+1)}\right);
\end{eqnarray}
\STATE (S9) The server sets 
$\mu_{ij}^{(t+1)} = \mu_{ij}^{(t)}$, 
for $\ell \notin \mathcal{S}_D^{(t)}$, 
$\ell \sim (i,j)$;
\STATE (S10) For each $\ell \in \mathcal{S}_D^{(t)}$, $\ell \sim (i,j)$, 
the server sends~$\mu_{ij}^{(t+1)}$
 to user $i$;
 \STATE (S11) Each user $i$, such that 
 $(i,j) \sim \ell$, 
 $\ell \in \mathcal{S}_D^{(t)}$, 
 performs the following update:
 	\begin{eqnarray}
 	\label{eqn-update-hat-mu-ij}
	&\,& \widehat{\mu}_{ij}^{(t+1)} = \mu_{ij}^{(t+1)}\\ 
	&-& \nonumber \nu \,\rho\,\left( 
x_i^{(t+1)}-x_j^{(t+1)}-z_{ij}^{(t+1)}\right);
\end{eqnarray}
\ENDFOR
\end{algorithmic}
\end{center}
\end{algorithm}

 Functions $B_i(\cdot,\cdot)$ in \eqref{eqn-x-i-update-PDMM-1}
 and $B_{ij}(\cdot,\cdot)$ in \eqref{eqn-z-ij-update-PDMM-1} are instances of Bregman divergence, e.g., \cite{TomLuoPDMM}; for example, they can be taken as $B(u,v) = \frac{1}{2}\|u-v\|^2$. 
 The choice of functions $B_i(\cdot,\cdot)$
  and $B_{ij}(\cdot,\cdot)$ also 
  affect the computational cost of updates 
 \eqref{eqn-x-i-update-PDMM-1} and \eqref{eqn-z-ij-update-PDMM-1}, respectively. For example, for 
 $B_{ij}(u,v) = \frac{1}{2}\|u-v\|^2$, 
  update~\eqref{eqn-z-ij-update-PDMM-1}
   corresponds to evaluating a proximal operator of the 2-norm that is done via block soft-thresholding. 
    See also subsection 2.1 in~\cite{TomLuoPDMM}
     for the choices of $B_i(\cdot,\cdot)$
      that make update \eqref{eqn-x-i-update-PDMM-1} computationally cheap. 
      The positive parameters $\eta_i$ in 
      \eqref{eqn-x-i-update-PDMM-1} and 
      $\eta_{ij}$ in \eqref{eqn-z-ij-update-PDMM-1}
       weigh the Bregman divergence terms; the larger $\eta_i$ is, the closer 
       $x_{i}^{(t+1)}$ is to $x_i^{(t)}$, 
       i.e., the smaller steps the algorithm makes.
        A similar effect is achieved with $\eta_{ij}$ in \eqref{eqn-z-ij-update-PDMM-1}. 
          Quantity $\widehat{\mu}_{ij}^{(t)}$ 
          in \eqref{eqn-update-hat-mu-ij}
        is an auxiliary dual variable associated with the dual variable ${\mu}_{ij}^{(t)}$. 
         The update step \eqref{eqn-update-hat-mu-ij}
          is a backward dual step 
          that is introduced for improving the algorithm's stability that may otherwise be violated due to the parallel and randomized nature of 
          primal variable updates; 
          see~\cite{TomLuoPDMM} for details. 
          Similarly, parameters $\tau>0$ and $\nu>0$ in \eqref{eqn-mu-ij-PDMM} and \eqref{eqn-update-hat-mu-ij},  respectively, 
           are ``damping'' factors in dual variable updates, that are again used to stabilize the algorithm trajectory.
          
With Algorithm~\ref{alg-PDMM-FL}'s initialization, we can set 
the $z_{ij}^{(0)}$'s, 
$\mu_{ij}^{(0)}$'s, and 
 and $\widehat{\mu}_{ij}^{(0)}$'s arbitrarily. For example, they can be all set to zero. 
  For the initialization of the $x_i$'s, 
  we need that $x_j^{(0)}$, 
  $j \neq i$, is available at each user $i$.
  This can be achieved by, e.g., setting 
  $x_i^{(0)}=0$, for all $i$, 
  or by letting the server send a common initial point  
  $y^{(0)} \in {\mathbb R}^d$ to all users prior to the algorithm start, so that each user $i$ sets $x_i^{(0)}=y^{(0)}$. The initial point~$y^{(0)}$  may also be obtained by (approximately) solving \eqref{eq:std_fed} via a (non-personalized, standard) FL algorithm, e.g., FedAvg.

With Algorithm~\ref{alg-PDMM-FL}, 
the server maintains and updates the $N(N-1)$ $d \times 1$-sized dual variables 
 $\mu_{ij}^{(t+1)}$, $i,j=1,\ldots,N$, $i \neq j$. Each user $i$ maintains and updates the 
 $d \times 1$-sized primal variable 
 $x_i^{(t)}$; $N-1$ $d\times 1$-sized primal variables $z_{ij}^{(t)}$, $j \neq i$; 
 and $N-1$ $d \times 1$-sized auxiliary dual variables $\widehat{\mu}_{ij}^{(t)}$, 
 $j \neq i$.
  
With Algorithm~\ref{alg-PDMM-FL}, 
communication from users to the server (``uplink'') takes place at steps (S5) and (S6). Note that, at each $t$, during steps (S5) and (S6), the server receives exactly $S_P$ $d \times 1$-sized (real vectors)  messages. Here, $S_P$ is the design parameter that can be taken to be much smaller than $N$, hence 
the uplink communication does not incur high overhead. Communication from the server to users (``downlink'') takes place at steps (S7) and (S10). At step (S7), 
the server brodcasts $S_D$ $d \times 1$-sized messages. At step (S10), the server transmits 
a total of $S_D$ $d \times 1$-sized messages 
to different users. (Specifically, 
variable $\mu_{ij}^{(t+1)}$ is sent to 
user $i$, for $\ell \sim (i,j)$, $\ell \in {\mathcal S}_D^{(t)}$.) Therefore, the downlink communication involves a total of $(S_P+S_D)$ $d \times 1$-sized messages per iteration~$t$. Quantity $S_D$ is also a design parameter that can be set to be much smaller than $N$; hence, 
the downlink communication does not incur 
a significant communication overhead.

By applying Theorem~3 in \cite{TomLuoPDMM} it can be shown  that, under appropriately chosen tuning parameters, 
$\mathbb{E}\left[ 
F(x^{(t)}) - F^\star\right]
= O(1/t)$, 
where we recall that $F$ is the objective function defined in \eqref{eq:per_fed_cc}, $F^\star = \inf_x F(x)$, 
and $x^{(t)}
= ((x_1^{(t)})^\top,\ldots,(x_N^{(t)})^\top)^\top$ is generated by Algorithm~\ref{alg-PDMM-FL}. 

\section{Numerical results}\label{sec:numerical_results}

We now present numerical simulations. In the first set of experiments, we evaluate the cluster recovery abilities, as well as personalization and generalization abilities of the proposed formulation (\ref{eq:per_fed_cc}) and compare it with alternatives. 

\begin{figure}[h]
    \centering
    \includegraphics[width=\columnwidth]{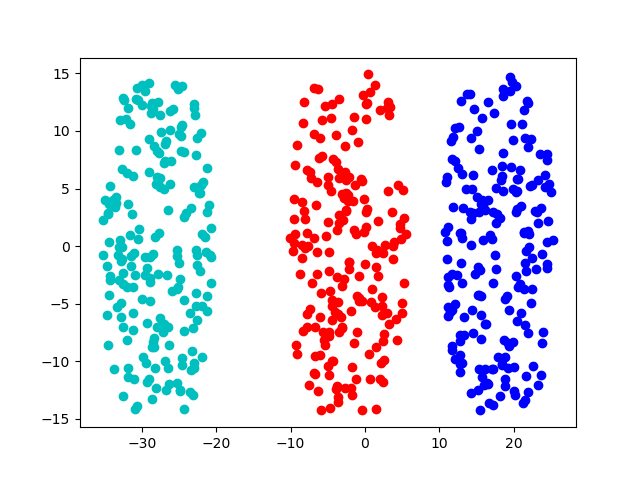}
    \caption{Training data, generated by the process described in Section \ref{sec:numerical_results}. The dashed lines represent optimal separators of the ellipses, within a cluster.}
    \label{fig:example_data}
\end{figure}

We now describe the first set of experiments. We consider a supervized binary classification problem. The generated data contains $K=3$ clusters. For each cluster, for each given label/class ($\pm 1$), the data comes from a uniform distribution over an ellipse in $\mathbb{R}^2$. The two ellipses that correspond to different classes for a given cluster overlap, so that the data in each cluster is not linearly separable. We generate 200 (training) data points from the distributions from each cluster, 100 points per class. Then, we associate to each cluster 20 FL users. Each FL user samples 10 data points out of the 200 data points available in its cluster. Hence, the data for all users within a cluster comes from the same distribution. Figure \ref{fig:example_data} illustrates the data, where different colors corresponds to different clusters, while the dashed lines represent optimal separators for each cluster, computed using the squared Hinge loss, i.e. the separators that minimize the squared Hinge loss over the full training data, for each cluster. The squared Hinge loss is used throughout the simulations, as the local loss function of each user $i$, and is given by
\begin{equation*}
    f_i(x) = \frac{c\|w\|^2}{2} + \frac{1}{m}\sum_{j = 1}^m \max \big\{0, 1 - \ell_{ij}(\langle w,a_{ij}\rangle - b) \big\}^2, 
\end{equation*} where $m$ represents the number of (local) data points, $(a_{ij},\ell_{ij})_{j = 1}^m$, represent the data points and class labels at user $i$, $c > 0$ is a penalty parameter that controls the regularization, while $x = \begin{bmatrix} w, b \end{bmatrix} \in \mathbb{R}^{d + 1}, \: d = 2$, represents the vector that defines the classifier. That is, the classifier based on vector $x= \begin{bmatrix} w, b \end{bmatrix}$ takes a feature vector $a \in \mathbb{R}^2$ as input and predicts its label as $\ell = \text{sign} (\langle w, a\rangle - b)$. Throughout the experiments, we set the parameter $c = 10^{-3}$, in order to put more weight on the classification performance of the method.

We compare the proposed formulation (\ref{eq:per_fed_cc}) with the alternatives in (\ref{eq:std_fed}), (\ref{eq:per_fed}) and (\ref{eq:per_fed_richtarik}), referred to here as the global model, local models, and squared penalty, respectively. Formulation (\ref{eq:std_fed}) corresponds to a standard, non-personalized FL solution. That is, the classifier vectors $x_i$'s with (\ref{eq:std_fed}) are equal for all users, i.e., $x_i = y^*$, where we recall that $y^*$ is the solution to (\ref{eq:std_fed}). With formulation (\ref{eq:per_fed}), each user $i$'s classifier vector equals $x_i = y_i^* = \argmin f_i(x)$. In addition, we compare the proposed formulation (\ref{eq:per_fed_cc}) with an oracle model that knows beforehand the clustering structure of the users; then, for each user $i$ within a cluster $C_k$, the oracle lets user $i$'s classifier vector be $x_i= \argmin_{x} \sum_{j \in C_k} f_{k}(x)$ We expect that the oracle model performs best in the considered setup among all methods, as it has an unfair advantage of knowing the cluster structure beforehand, and the data distributions of different clusters are very different, so data from a different cluster confuses another cluster's classifier. To evaluate solutions (\ref{eq:std_fed})-(\ref{eq:per_fed_richtarik}), we used CVXPY \cite{diamond2016cvxpy}, \cite{agrawal2018rewriting}. 

\begin{figure}[h]
    \centering
    \includegraphics[width=\columnwidth]{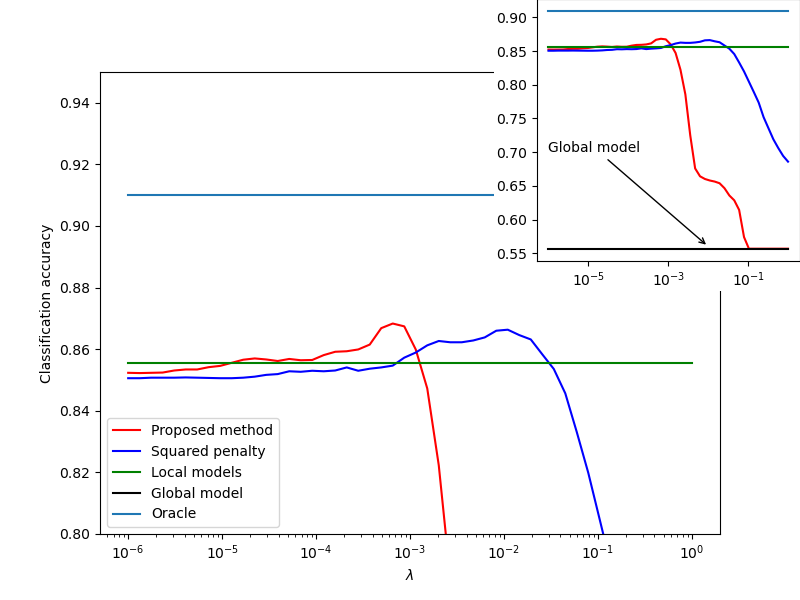}
    \caption{The average classification accuracy across all users and all clusters. We can see that the two personalization methods achieve both personalization, as they outperform the global model, as well as generalization, as they outperform the strictly local model, for certain values of $\lambda$.}
    \label{fig:results}
\end{figure}

\begin{figure}[h]
    \centering
    \includegraphics[width=\columnwidth]{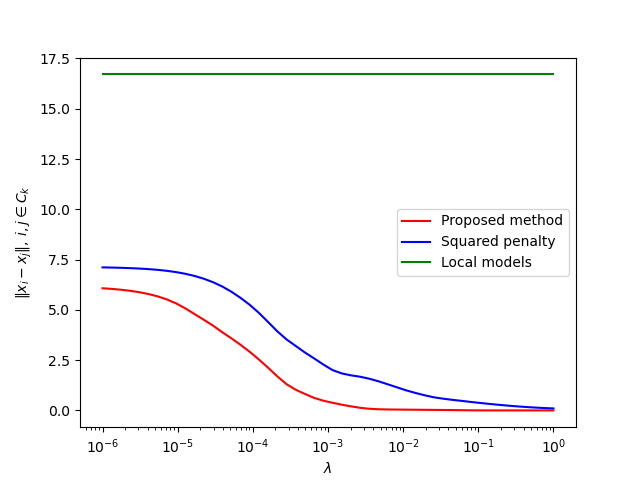}
    \caption{The average distance between models across all users and all clusters. We can see that distance between models increases, as $\lambda$ decreases, which is to be expected, as the models fit better to their local data. However, the proposed method results in more compact solutions within clusters, compared to the squared penalty one.}
    \label{fig:dist}
\end{figure}

In order to evaluate generalization and personalization abilities of different methods, we evaluate testing accuracy of the corresponding classifiers with respect to a newly generated test data. More precisely, let $x_i$ be a classifier vector for user $i$ obtained through training via any of the methods (\ref{eq:std_fed})-(\ref{eq:per_fed_richtarik}). For each user $i$, we then evaluate the testing accuracy of the classifier $x_i$ with respect to the full testing data set for the cluster to which user $i$ belongs to. We then average the testing accuracy across all users $i=1,\ldots,N$. For each cluster, the testing data is generated by drawing new samples (new with respect to training data) from the same distributions according to which the training data is generated. Methods (\ref{eq:per_fed_cc}) or (\ref{eq:per_fed_richtarik}) then exhibit generalization if the average testing accuracy is above the average testing accuracy of local models; they exhibit personalization if their average testing accuracy is above that of the global model. The performance of the models, for different values of $\lambda$, is presented in Figure \ref{fig:results}. Additionally, we present the average Euclidean distance between the classifier vectors $x_i$'s belonging to the same cluster. The results are summarized in Figure \ref{fig:dist}.

Figures \ref{fig:results} and \ref{fig:dist} show the following. For $\lambda$ sufficiently large, our method enforces consensus, and effectively performs as the standard FL method (\ref{eq:std_fed}). For $\lambda$ sufficiently small, the proposed method achieves both personalization, as it significantly outperforms the global model, as well as generalization, as the performance on the the complete cluster data is better than the strictly local models. Compared to the squared penalty model (\ref{eq:per_fed_richtarik}), we note that our method recovers the global model for sufficiently large $\lambda$, while the squared penalty method can recover the global model only asymptotically, as $\lambda$ tends to infinity. The highest average accuracy is achieved by our method, being at $86.8\%$, compared to the highest average accuracy of the squared penalty method, being at $86.6\%$. Figure \ref{fig:dist} shows that our method constantly produces more compact clusters, i.e. the average distance between solutions within clusters is constantly smaller than the one produced by the squared penalty method \footnote{Note that the considered simulation setup is such that the inter cluster generalization does not help. That is, data distributions across different clusters are so far apart, that the oracle, that ignores the data from other clusters, works best. Clearly, if the data distributions across different clusters are very close, another oracle that exploits inter-cluster generalization may be considered. Consider the extreme case when the data distributions across the three clusters are mutually (almost) equal, i.e., the distribution difference is negligible, 
but this is not known beforehand. Then, clearly, the global model (\ref{eq:std_fed}) becomes best-performing and an "oracle" model, in the sense that it implicitly utilizes the unknown cluster structure. Our formulation (\ref{eq:per_fed_cc}) can uncover this and actually match the global model for $\lambda$ above a threshold, and hence perform optimally. On the other hand, (\ref{eq:per_fed_richtarik}) gets closer to the global model as $\lambda$ increases, but never matches it. This illustration presents a scenario when there is clear merit in across-cluster generalization that the proposed formulation (\ref{eq:per_fed_cc}) is able to harness.}. 

Hence, we can see that the proposed formulation (\ref{eq:per_fed_cc}) achieves a comparable or slightly better peak accuracy with respect to (\ref{eq:per_fed_richtarik}), while producing more compact models, i.e., significantly reducing the number of distinct models that need to be kept in the overall FL system. We report that the proposed model (\ref{eq:per_fed_cc}) exactly recovers the cluster structure (produces equal user models within clusters and finds 3 clusters) for $\lambda \in (0.0892, 0.0919)$. 

Finally, we evaluated the performance of PDMM for solving (\ref{eq:per_fed_cc}). This result, as well as some additional numerical simulations, can be found in Appendix \ref{app:experiments}.

\section{Conclusion}\label{sec:conclusion}
We proposed a novel approach to personalized federated learning that, in addition to personalization and generalization, 
allows for clustering of users' local models. %
The approach is based on a novel formulation of personalized FL wherein we minimize the sum of local users' costs with respect to their local models, subject to a penalization term that penalizes the local models' differences via a sum-of-norms penalty. 
 We prove exact cluster recovery guarantees 
 for a general class of local users' costs, 
 assuming that the penalty parameter~$\lambda$ that weighs the sum-of-norms penalty falls within 
 an appropriately defined range.
 We further explicitly characterize this range in terms of within-clusters and across-clusters heterogeneity of
 local users' costs (models). As an interesting byproduct, these results 
 represent a direct generalization of convex clustering recovery guarantees for more general per-data point losses. Next, we propose an efficient algorithm based on the Parallel Direction Method of Multipliers (PDMM) to solve the proposed formulation in a federated server-users setting. Numerical experiments illustrate and corroborate the results.


\bibliographystyle{IEEEtran}
\bibliography{IEEEabrv,bibliography}

\begin{thebibliography}{10}
\providecommand{\url}[1]{#1}
\csname url@samestyle\endcsname
\providecommand{\newblock}{\relax}
\providecommand{\bibinfo}[2]{#2}
\providecommand{\BIBentrySTDinterwordspacing}{\spaceskip=0pt\relax}
\providecommand{\BIBentryALTinterwordstretchfactor}{4}
\providecommand{\BIBentryALTinterwordspacing}{\spaceskip=\fontdimen2\font plus
\BIBentryALTinterwordstretchfactor\fontdimen3\font minus
  \fontdimen4\font\relax}
\providecommand{\BIBforeignlanguage}[2]{{%
\expandafter\ifx\csname l@#1\endcsname\relax
\typeout{** WARNING: IEEEtran.bst: No hyphenation pattern has been}%
\typeout{** loaded for the language `#1'. Using the pattern for}%
\typeout{** the default language instead.}%
\else
\language=\csname l@#1\endcsname
\fi
#2}}
\providecommand{\BIBdecl}{\relax}
\BIBdecl

\bibitem{pmlr-v54-mcmahan17a}
\BIBentryALTinterwordspacing
B.~McMahan, E.~Moore, D.~Ramage, S.~Hampson, and B.~A.~y. Arcas,
  ``{Communication-Efficient Learning of Deep Networks from Decentralized
  Data},'' in \emph{Proceedings of the 20th International Conference on
  Artificial Intelligence and Statistics}, ser. Proceedings of Machine Learning
  Research, A.~Singh and J.~Zhu, Eds., vol.~54.\hskip 1em plus 0.5em minus
  0.4em\relax PMLR, 20--22 Apr 2017, pp. 1273--1282. [Online]. Available:
  \url{https://proceedings.mlr.press/v54/mcmahan17a.html}
\BIBentrySTDinterwordspacing

\bibitem{yu2020salvaging}
T.~Yu, E.~Bagdasaryan, and V.~Shmatikov, ``Salvaging federated learning by
  local adaptation,'' \emph{arXiv preprint arXiv:2002.04758}, 2020.

\bibitem{mocha}
V.~Smith, C.-K. Chiang, M.~Sanjabi, and A.~S. Talwalkar, ``Federated multi-task
  learning,'' in \emph{Advances in Neural Information Processing Systems},
  I.~Guyon, U.~V. Luxburg, S.~Bengio, H.~Wallach, R.~Fergus, S.~Vishwanathan,
  and R.~Garnett, Eds., vol.~30.\hskip 1em plus 0.5em minus 0.4em\relax Curran
  Associates, Inc., 2017.

\bibitem{richtarik}
F.~Hanzely and P.~Richtárik, ``Federated learning of a mixture of global and
  local models,'' 2021.

\bibitem{fallah_meta}
A.~Fallah, A.~Mokhtari, and A.~E. Ozdaglar, ``Personalized federated learning
  with theoretical guarantees: {A} model-agnostic meta-learning approach,'' in
  \emph{Advances in Neural Information Processing Systems 33: Annual Conference
  on Neural Information Processing Systems 2020, NeurIPS 2020, December 6-12,
  2020, virtual}, H.~Larochelle, M.~Ranzato, R.~Hadsell, M.~Balcan, and H.~Lin,
  Eds., 2020.

\bibitem{wang2019federated}
K.~Wang, R.~Mathews, C.~Kiddon, H.~Eichner, F.~Beaufays, and D.~Ramage,
  ``Federated evaluation of on-device personalization,'' 2019.

\bibitem{hinton}
G.~Hinton, O.~Vinyals, and J.~Dean, ``Distilling the knowledge in a neural
  network,'' \emph{arXiv preprint arXiv:1503.02531}, 2015.

\bibitem{zhang2018deep}
Y.~Zhang, T.~Xiang, T.~M. Hospedales, and H.~Lu, ``Deep mutual learning,'' in
  \emph{Proceedings of the IEEE Conference on Computer Vision and Pattern
  Recognition}, 2018, pp. 4320--4328.

\bibitem{Bistritz2020DistributedDF}
I.~Bistritz, A.~Mann, and N.~Bambos, ``Distributed distillation for on-device
  learning,'' in \emph{NeurIPS}, 2020.

\bibitem{joshi}
Y.~J. Cho, J.~Wang, T.~Chiruvolu, and G.~Joshi, ``Personalized federated
  learning for heterogeneous clients with clustered knowledge transfer,''
  \emph{ArXiv}, vol. abs/2109.08119, 2021.

\bibitem{ghosh2021efficient}
A.~Ghosh, J.~Chung, D.~Yin, and K.~Ramchandran, ``An efficient framework for
  clustered federated learning,'' 2021.

\bibitem{sattler}
F.~Sattler, K.-R. Müller, and W.~Samek, ``Clustered federated learning:
  Model-agnostic distributed multitask optimization under privacy
  constraints,'' \emph{IEEE Transactions on Neural Networks and Learning
  Systems}, vol.~32, no.~8, pp. 3710--3722, 2021.

\bibitem{Mansour20}
Y.~Mansour, M.~Mohri, J.~Ro, and A.~T. Suresh, ``Three approaches for
  personalization with applications to federated learning,'' \emph{ArXiv}, vol.
  https://arxiv.org/abs/2002.10619, 2020.

\bibitem{Sun21}
D.~Sun, K.-C. Toh, and Y.~Yuan, ``Convex clustering: Model, theoretical
  guarantee and efficient algorithm,'' \emph{Journal of Machine Learning
  Research}, vol.~22, 2021.

\bibitem{Hanzely20}
F.~Hanzely, S.~Hanzely, S.~Horvath, and P.~Richtarik, ``Lower bounds and
  optimal algorithms for personalized federated learning,'' in \emph{Advances
  in Neural Information Processing Systems 33 (NeurIPS 2020)}, 2020.

\bibitem{TomLuoPDMM}
H.~Wang, A.~Banerjee, and Z.-Q. Luo, ``Parallel direction method of
  multipliers,'' in \emph{Advances in Neural Information Processing Systems},
  Z.~Ghahramani, M.~Welling, C.~Cortes, N.~Lawrence, and K.~Q. Weinberger,
  Eds., vol.~27.\hskip 1em plus 0.5em minus 0.4em\relax Curran Associates,
  Inc., 2014.

\bibitem{BianchiRobust}
W.~Ben-Ameur, P.~Bianchi, and J.~Jakubowicz, ``Robust distributed consensus
  using total variation,'' \emph{IEEE Trans. Aut. Contr.}, vol.~61, no.~6,
  2016.

\bibitem{Panahi17}
A.~Panahi, D.~Dubhashi, F.~D. Johansson, and C.~Bhattacharyya, ``Clustering by
  sum of norms: Stochastic incremental algorithm, convergence and cluster
  recovery,'' in \emph{Proceedings of the 34th International Conference on
  Machine Learning, PMLR 70}, 2017, pp. 2769--2777.

\bibitem{TomLuo1}
M.~Hong, M.~Razaviyayn, Z.-Q. Luo, and J.-S. Pang, ``A unified algorithmic
  framework for block-structured optimization involving big data: {W}ith
  applications in machine learning and signal processing,'' \emph{IEEE Sig.
  Proc. Mag.}, vol.~33, no.~1, pp. 57--77, 2016.

\bibitem{diamond2016cvxpy}
S.~Diamond and S.~Boyd, ``{CVXPY}: {A} {P}ython-embedded modeling language for
  convex optimization,'' \emph{Journal of Machine Learning Research}, vol.~17,
  no.~83, pp. 1--5, 2016.

\bibitem{agrawal2018rewriting}
A.~Agrawal, R.~Verschueren, S.~Diamond, and S.~Boyd, ``A rewriting system for
  convex optimization problems,'' \emph{Journal of Control and Decision},
  vol.~5, no.~1, pp. 42--60, 2018.

\bibitem{ConvexClusteringClusterpath}
T.~D. Hocking, A.~Joulin, F.~Bach, and J.-P. Vert, ``{Clusterpath An Algorithm
  for Clustering using Convex Fusion Penalties},'' in \emph{{28th international
  conference on machine learning}}, United States, Jun. 2011, p.~1.

\end{thebibliography}


\appendices
\section{Practical considerations}\label{app:further_discuss}

In this section we discuss some practical aspects regarding the penalty parameter $\lambda$ of the formulation (\ref{eq:per_fed_cc}). Namely, we present upper and lower bounds alternative to (\ref{eq:condition-recovery})-(\ref{eq:condition-unique}), that still uncover the clustering structure. We also discuss how to select $\lambda$ in practice. 

\begin{remark} Note that, in \eqref{eq:condition-recovery}, 
the right hand side depends on $\lambda$ 
through the $w_k^\star(\lambda)$'s. 
We can replace condition~\eqref{eq:condition-recovery}  
 with a more conservative condition as follows. 
 Denote by 
 $G:\,{\mathbb R}^{N\,K} \rightarrow \mathbb R$, 
 the following function:
  $G(w_1,\ldots,w_K) = \frac{1}{N}\sum_{k=1}^K n_k \,g_k(w_k)$. By Assumption~\ref{assumption-f-i-s}, 
  $G$ is coercive and hence it has compact sub-level sets. Also, let $g^\star$ be the optimal value of~\eqref{eq:clustered-cc}, i.e.:
  \begin{equation}
      g^\star = \frac{1}{N}\sum_{k = 1}^Kn_kg_k(w_k^\star) + \lambda \sum_{k \neq l}n_kn_l \| w_k^\star - w_l^\star \|.
  \end{equation} Then, clearly, $G(w^\star_1,\ldots,w^\star_K) \leq g^\star$. On the other hand, we have
  \begin{equation}
      g^\star \leq \frac{1}{N}\sum_{k = 1}n_k g_k(y) + \sum_{l \neq k} n_k n_l \|y^\star - y^\star \| = \frac{1}{N}\sum_{i = 1}^Nf_i(y^\star) = f^\star,
  \end{equation} where we recall that $y^\star$ is a solution to $(\ref{eq:std_fed})$. Therefore, we conclude that, for any $\lambda \geq 0$,
  $\{w_k^\star(\lambda)\}$ belongs to the compact set
   \begin{equation}
   \label{eqn-compact-set-W}
   \mathcal{W}:=\{w_1,\ldots,w_K:\,G(w_1,\ldots,w_K) \leq 
   f^\star\}.
   \end{equation}
   Therefore, Theorem~\ref{lemma-recovery-cond} continues to hold if we replace \eqref{eq:condition-recovery} 
   with the following more stringent requirement: 
\begin{equation}
\label{eq:condition-recovery-stringent}
\lambda \geq \underline{\lambda}:=\max_{k=1,\ldots,K} \max_{i,j\in C_k}
\max_{\{w_k\} \in \mathcal{W}}\frac{ \|\nabla f_i(w_k) - \nabla f_j (w_k)\|}{n_k},    
\end{equation} 
Clearly, set $\mathcal{W}$ in \eqref{eq:condition-recovery-stringent} can be replaced with a larger compact set $\{w_1\ldots,w_K:\,G(w_1,\ldots,w_K) \leq 
   \frac{1}{N}\sum_{i=1}^N f_i(x^\bullet)\} $, 
   with arbitrary $x^\bullet \in {\mathbb R}^d$, e.g., $x^\bullet=0$. 
\end{remark}

\begin{remark} Note that, in \eqref{eq:condition-unique}, 
 the right hand side also depends on $\lambda$ 
 through the $w_k^\star(\lambda)$. We can replace  \eqref{eq:condition-unique}
  with the following more conservative condition:
\begin{equation}
\label{eq:condition-unique-stringent}
\lambda < \overline{\lambda}:=\frac{\min_{k,l \in [K], k\ne l} \min_{\{w_k\} \in \mathcal{W}} \|\nabla g_k(w_k) - \nabla g_l (w_k)\|}{2\max_{k \in [K]}\sum_{l \ne k}n_l},
\end{equation}
where set $\mathcal{W} \subset {\mathbb R}^{d\,K}$ is defined in~\eqref{eqn-compact-set-W}. 
\end{remark}

\begin{remark} Note that, if 
$\underline{\lambda} < \overline{\lambda}$, then, for any 
 $\lambda \in (\underline{\lambda}, \overline{\lambda})$,  
 both Theorems \ref{lemma-recovery-cond} and 
 \ref{lemma-recovery-uperb} hold, i.e., 
 formulation \eqref{eq:per_fed_cc}  
 perfectly recovers the $f_i$'s cluster structure  $C_1,\ldots,C_K$, and moreover the 
 models $w_k^\star$'s, $k=1,\ldots,K$, that correspond to different clusters, are mutually distinct. Intuitively, condition $\underline{\lambda} < \overline{\lambda}$ requires that the (appropriately scaled) within-clusters function heterogeneity 
 is smaller than the (appropriately scaled) 
between-clusters function heterogeneity.  
\end{remark}

In practice, we may not know quantities $\underline{\lambda}$ and 
 $\overline{\lambda}$. Similarly to 
 convex clustering approaches, e.g. \cite{ConvexClusteringClusterpath}, 
 we can solve~\eqref{eq:per_fed_cc} 
  for a set of values of the penalty parameter $\lambda$,  
  $\{\lambda^{(r)}\}$, $r=1,\ldots,R$, i.e., we can generate a solution path. In more detail, 
  we can set $\lambda^{(r+1)}=c \, \lambda^{(r)}$, $r=1,2,\ldots,$ for a small positive $\lambda^{(1)}$, where 
  $c>1$ is a constant, and $R$ is the smallest index $r$ such that the number of distinct vectors $x_i^\star(\lambda^{(R)})$, $i=1,\ldots,N$, is one.\footnote{Note that we know that, for 
  $\lambda \geq \widehat{\lambda}$, for some 
  $\widehat{\lambda}>0$, all the $x_i^\star(\lambda)$'s coincide~\cite{BianchiRobust}. Hence, 
  all the $x_i^\star(\lambda)$'s are necessarily mutually equal for 
  a certain $\lambda^{(R)}$, for some finite~$R>0$.}
   When solving~\eqref{eq:per_fed_cc} for $\lambda = \lambda^{(r+1)}$, 
   the numerical solver of~\eqref{eq:per_fed_cc}
   (e.g., see ahead Algorithm~\ref{alg-PDMM-FL}) can use warm start, i.e., it can be initialized with  $\{x_i^\star(\lambda^{(r)})\}$. 
      The resulting solution path 
     $\{x_i^\star(\lambda^{(r)})\}$, 
     $r=1,\ldots,R$, will typically have a non-increasing number of 
     mutually distinct models $x_i$'s,
      with $N$ distinct models for $\lambda^{(1)}$ and a sufficiently small 
      $\lambda^{(1)}$, and one distinct model 
      for~$\lambda^{(R)}$.
      
    If, for a certain value $r$ (or for a range of values $r$), we have that   
      $\lambda^{(r)} \in (\underline{\lambda}, \overline{\lambda})$ and there holds 
      $\underline{\lambda}<\overline{\lambda}$ (a hidden cluster structure exists), then 
      solution $\{x_i^\star(\lambda^{(r)})\}$
      satisfies Theorems \ref{lemma-recovery-cond} and \ref{lemma-recovery-uperb}, 
      i.e., the hidden cluster structure is uncovered. For a fixed $\lambda^{(r)}$, we 
      can check whether the hidden cluster structure is uncovered in an ``a posteriori'' way as follows.  
      We first fix the clustering $C_1=C_1(\lambda^{(r)}),\ldots,C_K=
      C_K(\lambda^{(r)})$  induced by  
       $\{x_i^\star(\lambda^{(r)})\}$, and then we check whether  
       conditions~\eqref{eq:condition-recovery-stringent} and 
       \eqref{eq:condition-unique-stringent} hold.  
       
      Even when we are not directly concerned with uncovering the hidden cluster structure in the sense of Theorems~\ref{lemma-recovery-cond} and \ref{lemma-recovery-uperb}, there are several  merits of computing the solution path, similarly to the convex clustering scenario,  e.g.,~\cite{ConvexClusteringClusterpath}.   %
        For example, depending on the requirements of a given application, we can select  the index $r$, i.e., the model 
        $\{x_i^\star(\lambda^{(r)})\}$ for which the number of distinct $x_i$'s 
        (closely) matches the need of the current application. Furthermore, increasing  
        $\lambda$ translates into improving the degree of generalization and reducing the degree of personalization, so one can select the appropriate $\{x_i^\star(\lambda^{(r)})\}$ according to 
        the current application needs.

\section{Further insights through a special case}\label{app:insight}

In this section, we provide further insight into the method, by analyzing a special case. We consider an explicit clustering structure, given by the following assumption.

\begin{assumption}\label{assumption-distance}
There exists a node partition $C_1, C_2,\ldots, C_K,$ and parameters $\epsilon_k, \delta_{kl} > 0, \: k,l = 1,\ldots,K, \: k\ne l,$ such that the following holds:
\begin{align}
    \sup_{x \in \R^d}\| \nabla f_i(x) - \nabla f_j(x) \| &\leq \epsilon_k, \: \forall i, j \in C_k, \label{eq:epsilon_sim} \\
    \inf_{x \in R^d}\| \nabla f_i(x) - \nabla f_j(x) \| &\geq \delta_{kl}, \: \forall i \in C_k, \: \forall j \in C_l, \: k \ne l. \label{eq:delta_diff}
\end{align}
\end{assumption} 

We then have the following result.
\begin{theorem}\label{lemma-lambda-bdds}
Let Assumptions \ref{assumption-f-i-s}, \ref{assumption-Lip-grad} and \ref{assumption-distance} hold. Then, for any $\lambda$ satisfying 
\begin{equation}\label{eq:lambda-bounds}
    \lambda \in \bigg[\max_{k \in [K]}\frac{\epsilon_k}{n_k},\frac{\min_{k\ne l} \big[\delta_{kl} - (\epsilon_k + \epsilon_l)\big]}{2\max_{k  \in [K]}\sum_{l\ne k}n_l}\bigg],
\end{equation} the conditions of Theorems \ref{lemma-recovery-cond} and \ref{lemma-recovery-uperb} are satisfied. 
\end{theorem}

\begin{remark} Note that conditions (\ref{eq:epsilon_sim}) and (\ref{eq:delta_diff}) can be interpreted as measures of within-cluster homogeneity and between-cluster heterogeneity, respectively. In particular, let $x_i^{\dagger} \in \mathbb{R}^d$ denote a optima of $f_i(x), \: i \in [N]$. Then, per (\ref{eq:epsilon_sim}) and (\ref{eq:delta_diff}), we have
\begin{align*}
    &\| \nabla f_j(x_i^{\dagger}) \| \leq \epsilon_k, \forall i,j \in C_k, \\
    &\| \nabla f_j(x_i^{\dagger}) \| \geq \delta_{kl}, \forall i \in C_k, \forall j \in C_l, k \neq l.
\end{align*} From convexity of $f_i$'s (Assumption \ref{assumption-f-i-s}) for any $k \in [K], \: i,j \in C_k$, if $\epsilon_k$ small, we can expect $f_j(x_i^{\dagger})$ to be a good approximation to $f_j(x_j^{\dagger})$. On the other hand, for any $k \neq l$ and $i \in C_k, \: j \in C_l$, using Lipschitz continuity of the gradients of $f_i$'s, we have 
\begin{equation*}
    f_j(x_i^{\dagger}) - f_j(x_j^{\dagger}) \geq \frac{1}{2L}\| \nabla f_j(x_i^{\dagger})\|^2 \geq \frac{\delta_{kl}^2}{2L}.
\end{equation*} Hence, for $\delta_{kl}$ large, $f_j(x_i^{\dagger})$ can be an arbitrarily bad approximation to $f_j(x_j^\dagger)$. Therefore, $\epsilon_k$ and $\delta_{kl}$ can be interpreted as natural measures of within-cluster homogeneity and between-cluster heterogeneity, respectively.
\end{remark}

\begin{remark} Theorem \ref{lemma-lambda-bdds} states that the clustering structure of the solution is maintained, for any choice of $\lambda$ satisfying (\ref{eq:lambda-bounds}). Compared to the results from Theorems \ref{lemma-recovery-cond} and \ref{lemma-recovery-uperb}, the resulting interval (if existent) is smaller, but the lower and upper bounds are independent of $\lambda$, and of the optimal solutions of problem (\ref{eq:clustered-cc}). 
\end{remark}

\begin{remark} Compared with (\ref{eq:condition-recovery-stringent}) and (\ref{eq:condition-unique-stringent}), while possibly smaller, the interval (\ref{eq:lambda-bounds}) provides a more natural interpretation: if the between-cluster heterogeneity is sufficiently larger than the within-cluster homogeneity, so the interval (\ref{eq:lambda-bounds}) is non-empty, a (strong) clustering structure among users exists, and can be recovered by (\ref{eq:per_fed_cc}).   
\end{remark}

We now prove Theorem \ref{lemma-lambda-bdds}.

\begin{proof}
From Assumption \ref{assumption-distance} it directly follows that 
\begin{equation*}
    \max_{k \in [K]} \max_{i,j\in C_k}\frac{ \|\nabla f_i(w_k^\star) - \nabla f_j (w_k^\star)\|}{n_k} \leq \max_{k \in [K]}\frac{\epsilon_k}{n_k}.
\end{equation*} Next, for any $k \in [K]$, and $i \in C_k$, we have
\begin{align*}
    \|\nabla g_k(x) - \nabla f_i(x) \| &= \bigg\|\frac{1}{n_k}\sum_{j \in C_k}\Big(\nabla f_j(x) - \nabla f_i(x)\Big)\bigg\| \\ &\leq \frac{1}{n_k}\sum_{j \in C_k \setminus \{i \}}\|\nabla f_j(x) - \nabla f_i(x)\| \\ &\leq \frac{n_k - 1}{n_k}\epsilon_k < \epsilon_k.
\end{align*} For any $k,l \in [K], \: k \ne l$, and any $i \in C_k, \: j \in C_l$, we then get
\begin{align*}
    \|\nabla g_k(x) &- \nabla g_l(x) \| \geq \|\nabla f_i(x) - \nabla f_j(x) \| \\  &- \|\nabla g_k(x) - \nabla f_i(x) \| - \|\nabla f_j(x) - \nabla g_l(x) \| \\ &> \delta_{kl} - \epsilon_k - \epsilon_l.
\end{align*} Plugging in $w^*_k$ and taking the $\min$ with respect to $k \neq l$ gives
\begin{equation*}
    \min_{k \neq l}\| \nabla g_k(w^*_k) - \nabla g_l(w^*_k) \| > \min_{k \neq l}(\delta_{kl} - \epsilon_k - \epsilon_l).
\end{equation*} Finally, dividing both sides by $2\max_{k \in [K]}\sum_{l \neq k}n_l$ gives the desired result.
\end{proof}

Moreover, assuming strong convexity, Assumption \ref{assumption-distance} implies a clustering structure among the local solutions, as shown by the following result.

\begin{theorem}\label{lemma-solutions-bounds}
Let each $f_i(x), \: i = 1,\ldots,N,$ be $\mu$-strongly convex, and let Assumptions \ref{assumption-Lip-grad}, \ref{assumption-distance} hold. Denote by $x_i^\dagger$ the (global) optima of $f_i(x), \: i = 1,\ldots,N$. Then, the following holds:
\begin{align}
    \|x_i^\dagger - x_j^\dagger\| &\leq \frac{\epsilon_k}{\mu}, \forall i,j \in C_k, \label{eq:upper-bdd} \\
    \|x_i^\dagger - x_j^\dagger\| &\geq \frac{\delta_{kl}}{L}, \forall i \in C_k, \: \forall j \in C_l, \: k \ne l. \label{eq:lower-bdd} 
\end{align}
\end{theorem}

\begin{remark} Theorem \ref{lemma-solutions-bounds} shows that optimal models of users belonging to the same clusters are at least $\tilde{\epsilon}$ close, with optimal models of users belonging to different clusters being at least $\tilde{\delta}$ apart. Here, $\tilde{\epsilon} = \max_{k \in [K]}\frac{\epsilon_k}{\mu}$, and $\tilde{\delta} = \min_{k\ne l}\frac{\delta_{kl}}{L}$. In the case that $\tilde{\epsilon} < \tilde{\delta}$, the clustering structure implied by Theorem \ref{lemma-solutions-bounds} is strong, in the sense that the local optima corresponding to different clusters are well separated.
\end{remark}

\begin{remark} In the case that $\tilde{\epsilon} < \tilde{\delta}$ and the interval given by (\ref{eq:lambda-bounds}) is non-empty, Theorems \ref{lemma-lambda-bdds} and \ref{lemma-solutions-bounds} show that a natural clustering structure among the user's costs exists. In addition, the proposed formulation (\ref{eq:per_fed_cc}) uncovers the said structure, for any $\lambda$ in a range that is independent of the optimal solutions of problem (\ref{eq:clustered-cc}).
\end{remark}

We now prove Theorem \ref{lemma-solutions-bounds}

\begin{proof}
From Assumption \ref{assumption-distance}, for $i \in C_k$, $j \in C_l$, $k \ne l$, and for all $x \in \mathbb{R}^d$, we have
\begin{align*}
    \delta_{kl} \leq \|\nabla f_i(x) - \nabla f_j(x)\|.
\end{align*} In particular, for $x = x_i^\dagger$, we have
\begin{align*}
    \delta_{kl} \leq \|\nabla f_j(x_i^\dagger)\| = \|\nabla f_j(x_i^\dagger) - \nabla f_j(x_j^\dagger)\| \leq L\|x_i^\dagger - x_j^\dagger\|,
\end{align*} implying (\ref{eq:lower-bdd}). Similarly, for $i,j \in C_k$, and for all $x \in \mathbb{R}^d$, we have
\begin{align*}
    \|\nabla f_i(x) - \nabla f_j(x) \| \leq \epsilon_k.
\end{align*} In particular, for $x = x_i^\dagger$, we have
\begin{align}\label{eq:intermed}
    \epsilon_k^2 \geq \|\nabla f_j(x_i^\dagger)\|^2 \geq 2\mu (f_j(x_i^\dagger) - f_j(x_j^\dagger)), 
\end{align} where we used the Polyak-Lojasiewitz inequality in the second step. From strong convexity of $f_j$, we have
\begin{align*}
    f_j(x_i^\dagger) &\geq f_j(x_j^\dagger) + \frac{\mu}{2}\|x_i^\dagger - x_j^\dagger\|^2.
\end{align*} Rearranging and plugging into (\ref{eq:intermed}), we get
\begin{equation*}
    \epsilon_k^2 \geq \mu^2\|x_i^\dagger - x_j^\dagger\|^2,
\end{equation*} which implies (\ref{eq:upper-bdd}).
\end{proof}

\section{Additional experiments}\label{app:experiments}

\begin{figure}[h]
    \centering
    \includegraphics[width=\columnwidth]{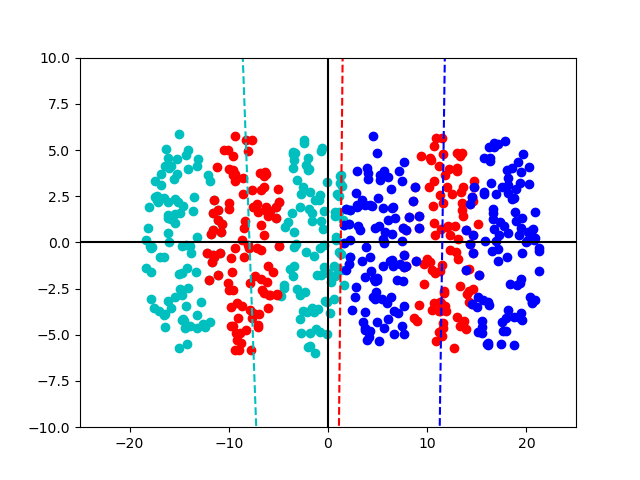}
    \caption{Data used for evaluating clustering recovery. Each ellipse contains 100 datapoints.}
    \label{fig:clust_data}
\end{figure}

\begin{figure}[h]
\centering
\begin{tabular}{ll}
\includegraphics[scale=0.18]{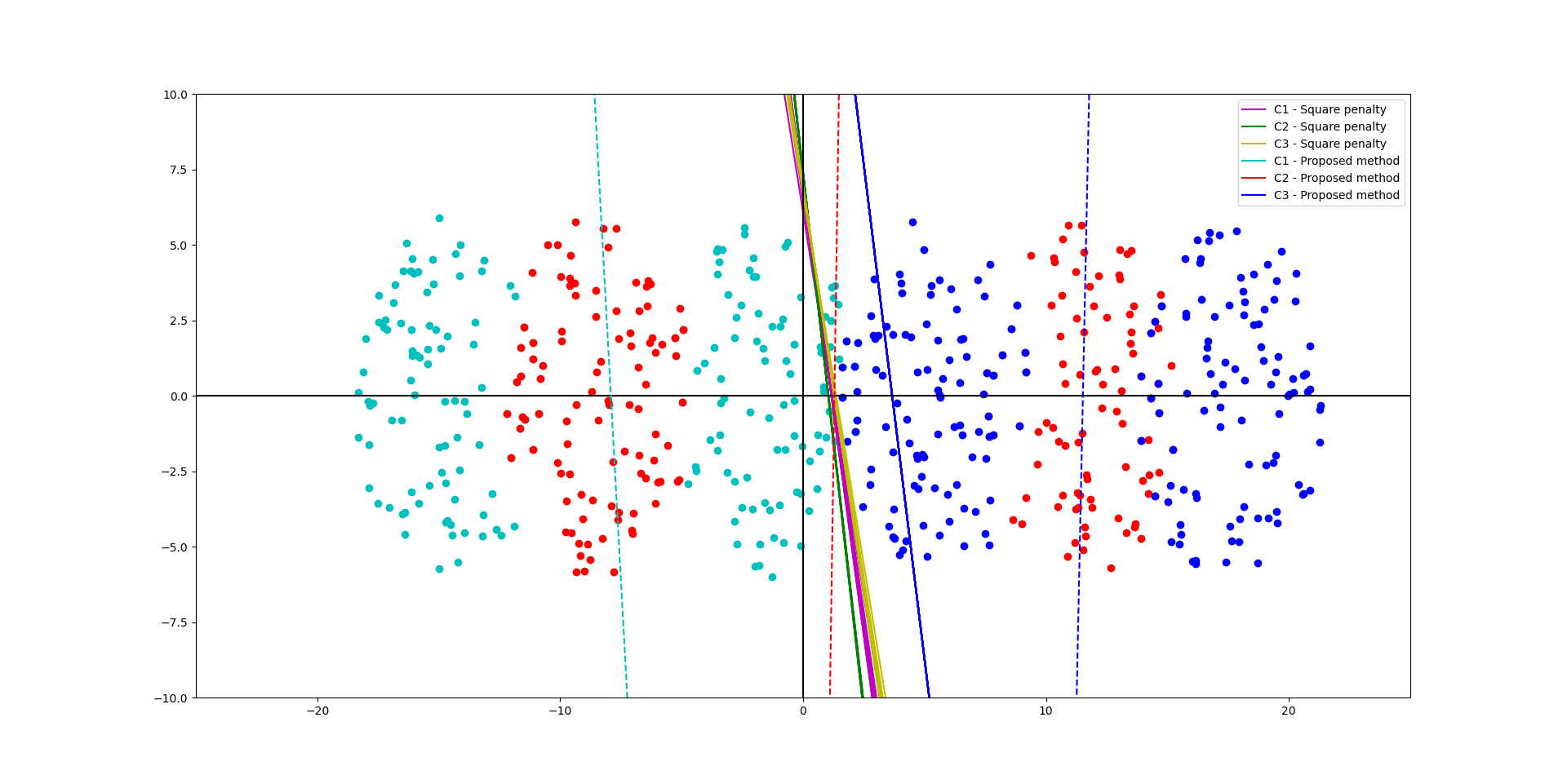}
&
\includegraphics[scale=0.18]{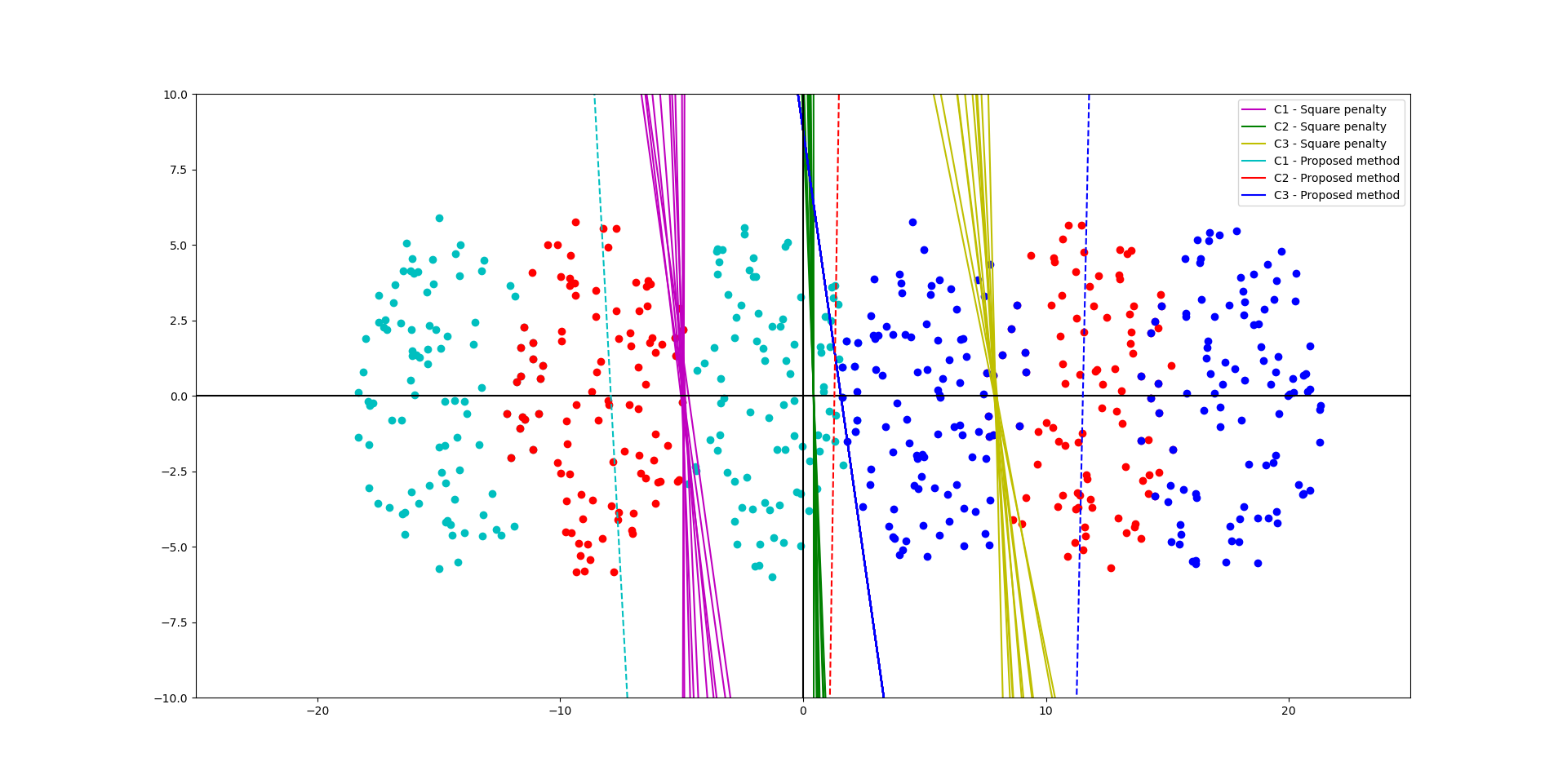}
\\
\includegraphics[scale=0.18]{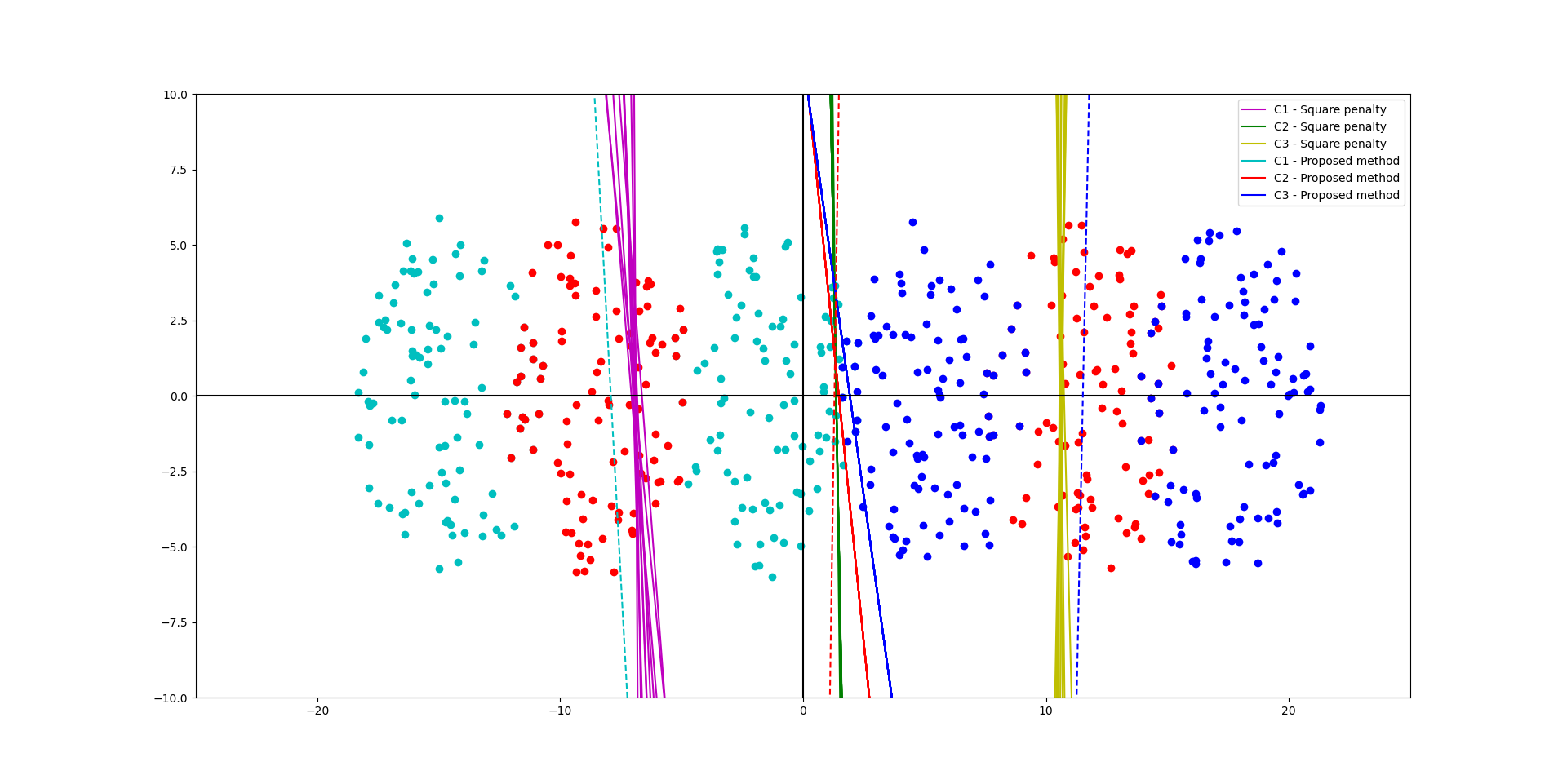}
&
\includegraphics[scale=0.18]{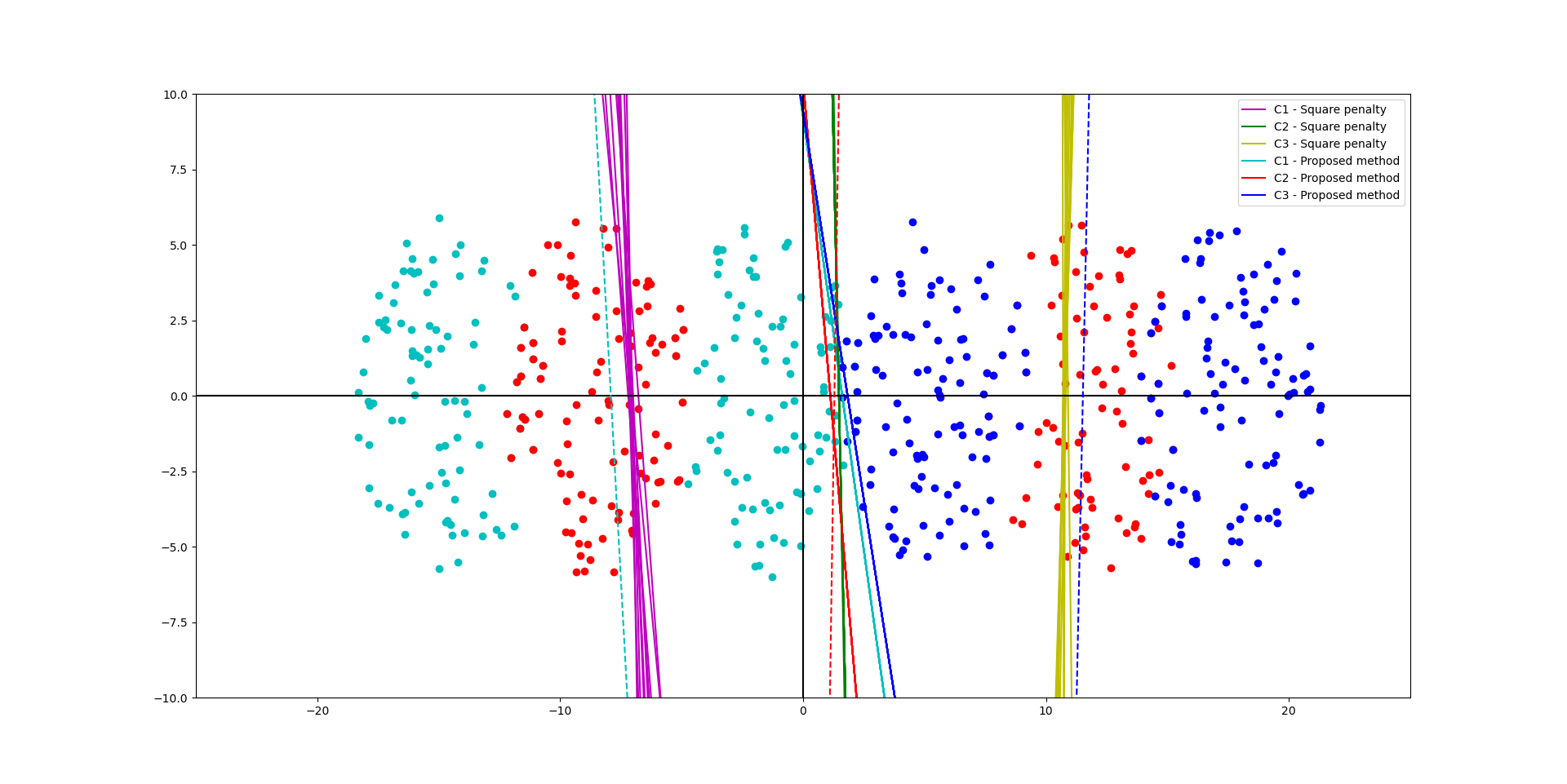}
\\
\includegraphics[scale=0.18]{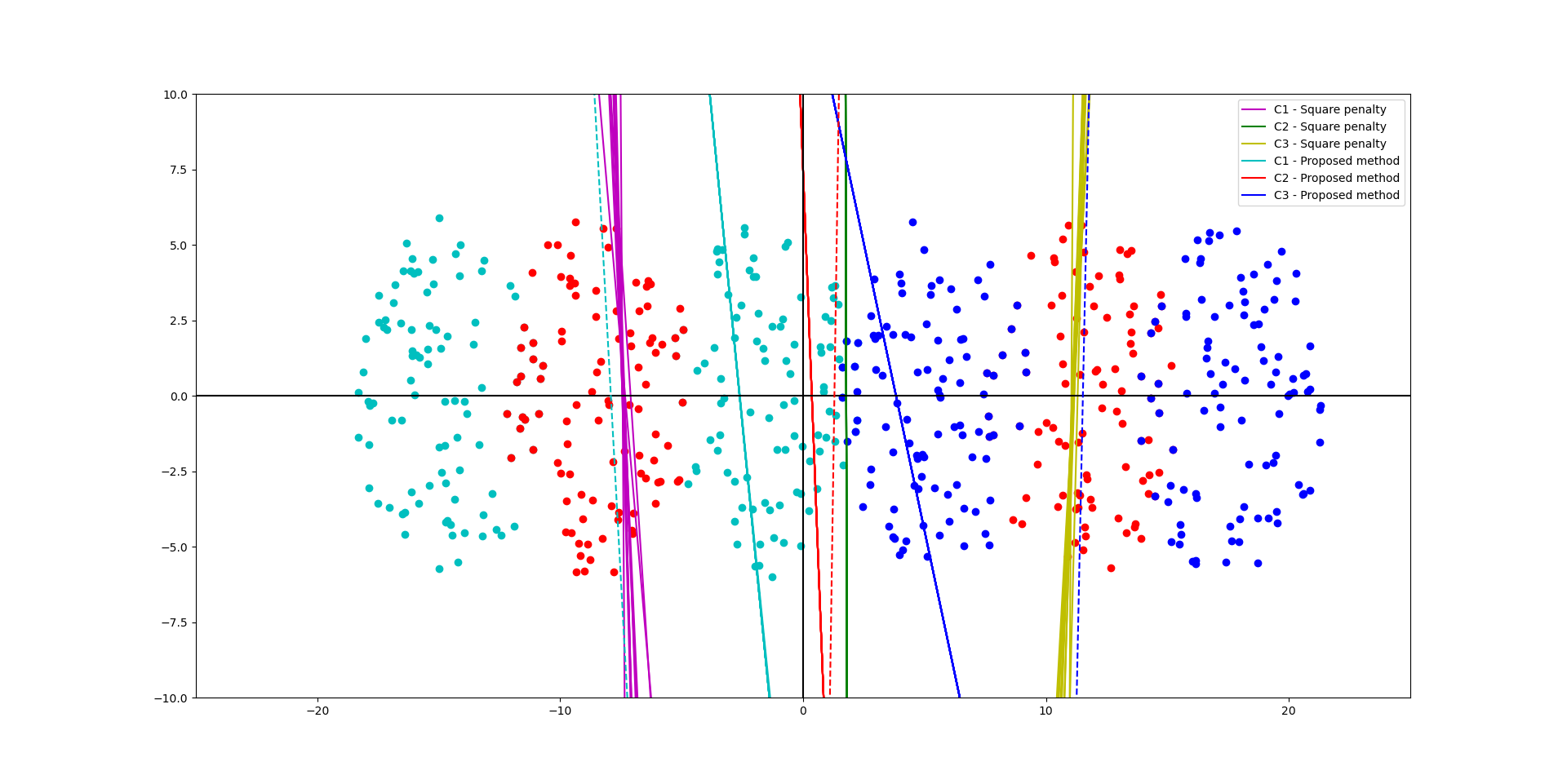}
&
\includegraphics[scale=0.18]{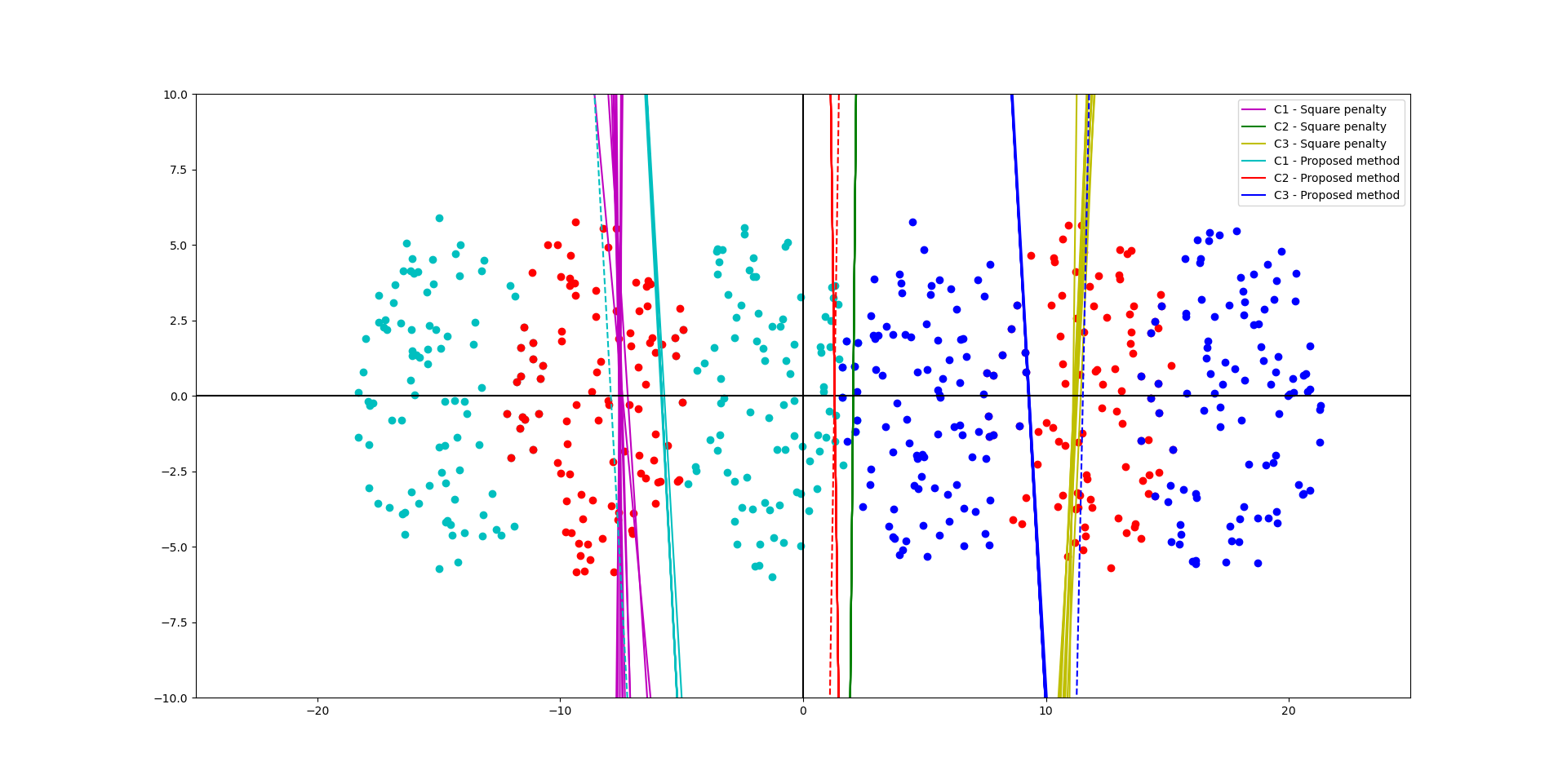}
\end{tabular}
\caption{Clustering structure of the models. The plots show how the number of distinct models changes with the penalty parameter $\lambda$. In particular, the values of $\lambda$ correspond to $\{1000, 1, 0.12, 0.08, 0.05, 0.03, 0.02, 0.01 \}$ from left to right, top to bottom, respectively. Note that the squared penalty method does not achieve consensus even for very high values of $\lambda$.}
\label{fig:clust_results}
\end{figure}

Here, we present some additional experiments. First, we evaluate the clustering recovery of (\ref{eq:per_fed_cc}) and compare it with (\ref{eq:per_fed_richtarik}). The data is generated using the same methodology described in Section \ref{sec:numerical_results}, with each ellipse containing $100$ data points, hence each cluster contains a total of $200$ data points. The dataset is shown in Figure \ref{fig:clust_data}. We generate $n = 10$ users per cluster, and each user samples $85\%$ of points from the cluster it belongs to, ensuring the resulting models will be similar for users within clusters. The performance of (\ref{eq:per_fed_cc}) and (\ref{eq:per_fed_richtarik}) is evaluated for different values of the penalty parameter $\lambda$, belonging to $\{1000, 1, 0.12, 0.08, 0.05, 0.03, 0.02, 0.01 \}$. The dashed lines in the figures correspond to optimal separators for each cluster, evaluated on the full data. The full lines represent model estimates corresponding to (\ref{eq:per_fed_cc}) and (\ref{eq:per_fed_richtarik}). The results are presented in Figure \ref{fig:clust_results}.

We can see that Figure \ref{fig:clust_results} corroborates the results from Theorems \ref{lemma-recovery-cond} and \ref{lemma-recovery-uperb}. In particular, for $\lambda$ sufficiently large, our method produces only one model across all users. For $\lambda$ higher than, but close to the theoretical upper bound from Theorem \ref{lemma-recovery-uperb}, the method produces two models across all the users (second row, left image). Finally, for $\lambda$ within the theoretical range, our method produces exactly three models, uncovering the clustering structure. We remark that, while the produced models are not optimal for the training data, they can still be used as a guideline: if a clustering solution is obtained, and the parameter $\lambda$ falls within the theoretical bounds suggested by Theorems \ref{lemma-recovery-cond} and \ref{lemma-recovery-uperb}, an innate clustering structure is uncovered, and the users that are selected in the clusters can focus on training their models within the cluster of similar users. Moreover, the left image in the third row suggests that the bounds obtained by Theorems \ref{lemma-recovery-cond} and \ref{lemma-recovery-uperb} are somewhat conservative - the proposed method produced 3 models across users, for $\lambda$ that is lower than the theoretical lower bound, meaning that in practice, the clustering structure can be recovered for a wider range of $\lambda$ than predicted by theory.    

\begin{figure}[h]
    \centering
    \includegraphics[width=\columnwidth]{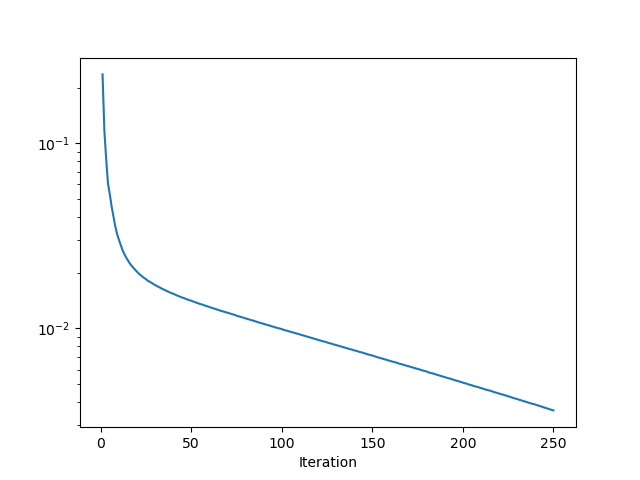}
    \caption{PDMM convergence for solving (\ref{eq:per_fed_cc}) in federated settings.}
    \label{fig:PDMM}
\end{figure}

In the second experiment, we present the convergence results of PDMM for solving the proposed method in federated settings. The dataset and the parameters are the same as was described in the setting above. For PDMM in the federated settings, in each iteration we randomly select a subset of users, being $40\%$ of the total users. The PDMM tuning parameters are selected as $\rho = 10, \: \eta = 10, \: \tau = \frac{4}{5}, \: \nu = \frac{1}{5}$. The Bregman divergence in PDMM is set to be the squared Euclidean distance. All variables in PDMM (primal and dual) are initialized to zero vectors. 

We evaluate the cost function in (\ref{eq:per_fed_cc}) achieved by PDMM at each iteration $t$. We also evaluate the optimal value of the cost in (\ref{eq:per_fed_cc}) via CVXPY. We then plot the difference (optimality gap) along iterations. To evaluate the performance of PDMM for (\ref{eq:per_fed_cc}), we used the primal variables $x_i$'s. In particular, we plot $F(x^{(t)}) - F^*$, where $F$ is the objective defined in (\ref{eq:per_fed_cc}) and $F^*$ denotes the optimal value of $F$, computed via CVXPY. The parameter $\lambda$ is set to $0.02.$ The results are presented in Figure \ref{fig:PDMM}.

While this is not pursued here, we remark that it is possible to reduce Algorithm~\ref{alg-PDMM-FL} 
complexity by dynamically (over iterations) exploiting the clustered structure of a solution to~\eqref{eq:per_fed_cc}. 
Namely, note that, for a solution $\{x_i^\star\}$ of \eqref{eq:per_fed_cc}, if $x_i^\star=x_j^\star$ for some $i \neq j$, then  $z_{ij}^\star=0$.  This may be exploited in the algorithm implementation as follows: once user $i$ detects that $z_{ij}^{(t)}$ is close to zero over a range of consecutive iterations, it can cease updating $z_{ij}^{(t)}$ and cease communicating it to the server, and it can also replace locally   $x_j^{(t)}$ with $x_i^{(t)}$ in its subsequent updates. This also translates into reducing the communication cost at the server side, as 
$x_j^{(t)}$ no longer needs to be communicated to user~$i$. In this way, storage, computational, and communication costs may be dynamically reduced as the algorithm evolves.

\onecolumn
\pagebreak 

\end{document}